\documentclass[preprint,12pt]{elsarticle}

\usepackage{lineno,hyperref}
\usepackage{natbib}
\usepackage{geometry}
\usepackage{fleqn}
\usepackage{graphicx}
\usepackage{newtxtext,newtxmath}
\usepackage{hyperref}
\usepackage{xurl}

\hypersetup{
    colorlinks,
    linkcolor={red!50!black},
    citecolor={blue!50!black},
    urlcolor={blue!80!black},
    pdfauthor={Lesego E. Moloko} 
}

\usepackage{booktabs}
\usepackage[justification=centering]{caption}
\modulolinenumbers[5]

\usepackage{framed}  
\usepackage{nomencl} 
\makenomenclature

\usepackage{color,soul}
\usepackage{subcaption}

\usepackage{bm}
\usepackage{amsmath}

\usepackage[detect-all]{siunitx} 

\graphicspath{{./figs/}}
\journal{Annals of Nuclear Energy}

\usepackage{multicol}
\renewcommand*{\nompreamble}{\begin{multicols}{2}}
\renewcommand*{\nompostamble}{\end{multicols}}
\usepackage{easyReview}
\usepackage{totcount}
\newtotcounter{TODO}

\begin{document}

\begin{frontmatter}

\title{Clustering and Uncertainty Analysis to Improve the Machine Learning-based Predictions of SAFARI-1 Control Follower Assembly Axial Neutron Flux Profiles}

\author[NESCA,NCSU]{Lesego E. Moloko\corref{correspondingauthor}}
\ead{lesego.moloko@necsa.co.za}
\author[NESCA,NCSU]{Pavel M. Bokov}
\author[NCSU]{Xu Wu}
\author[NCSU]{Kostadin N. Ivanov}

\cortext[correspondingauthor]{Corresponding author}

\address[NESCA]{The South African Nuclear Energy Corporation SOC Ltd (Necsa) \\
	Building-1900, P.O. Box 582, Pretoria 0001, South Africa \\}

\address[NCSU]{Department of Nuclear Engineering, North Carolina State University    \\
	Burlington Engineering Laboratories, 2500 Stinson Drive, Raleigh, NC 27695 \\}

\begin{abstract}
  The goal of this work is to develop accurate Machine Learning (ML) models for predicting the assembly axial neutron flux profiles in the SAFARI-1 research reactor, trained by measurement data from historical cycles. The data-driven nature of ML models makes them susceptible to uncertainties which are introduced by sources such as noise in training data, incomplete coverage of the domain, extrapolation and imperfect model architectures. To this end, we also aim at quantifying the approximation uncertainties of the ML model predictions. Previous work using Deep Neural Networks (DNNs) has been successful for fuel assemblies in SAFARI-1, however, not as accurate for control follower assemblies. The aim of this work is to improve the ML models for the control assemblies by a combination of supervised and unsupervised ML algorithms. The $k$-means and Affinity Propagation unsupervised ML algorithms are employed to identify clusters in the set of the measured axial neutron flux profiles. Then, regression-based supervised ML models using DNN (with prediction uncertainties quantified with Monte Carlo dropout) and Gaussian Process (GP) are trained for different clusters and the prediction uncertainty is estimated. It was found that applying the proposed procedure improves the prediction accuracy for the control assemblies and reduces the prediction uncertainty. Flux shapes predicted by DNN and GP are very close, and the overall accuracy became comparable to the fuel assemblies. The prediction uncertainty is however smaller for GP models.
\end{abstract}


\end{frontmatter}




\begin{table*}[!t]
\begin{framed}
\footnotesize
  \section*{Nomenclature}
    \begin{multicols}{2}
     \subsection*{Symbols}
        \begin{description}
        \itemsep-0.2em\small
        \item[\rm $\bm{b}$] vector of learnable biases
        \item[\rm $\mathcal{C}$] cluster
        \item[\rm $\mathbb{E}$] mathematical expectation
       \item[\rm $\mathcal{GP}$] Gaussian Process
        \item[\rm $\bm{h}^{(\ell)}$] vector of outputs from layer $\ell$
        \item[\rm $i$] sample index
        \item[\rm $I$] mutual information
        \item[\rm $\bm{I}$] identity matrix
        \item[\rm $H$] Shannon entropy
        \item[\rm $k$] number of clusters
        \item[\rm $\bm{K}$] cross-covariance matrix
        \item[\rm $l$] length-scale parameter
        \item[\rm $\ell$] layer index
        \item[\rm $L$] depth of a neural network
        \item[\rm $\mathcal{L}_{\text{dropout}}$] loss function in MCD 
        \item[\rm $m$] mean function in GP
        \item[\rm $N$] number of samples 
        \item[\rm $\mathcal{N}$] Gaussian (Normal) distribution
        \item[\rm $p$] dropout probability
        \item[\rm $\mathcal{P}$] set of clusters
        \item[\rm $\mathcal{S}$] set of objects
        \item[\rm $t$] forward pass index
        \item[\rm $T$] number of forward passes
       \item[$\bm{W}$] learnable weights
        \item[$\bm{x}$] input vector
        \item[$\bm{y}$] vector of target outputs
        \item[$\hat{\bm{y}}$] vector of predicted outputs
        \item[\rm $\bm{Z}_\ell$] dropout matrix before the $\ell$-th layer
        \item[\rm $\delta(\bm{x}_i,\bm{x}_j)$] Kronecker delta
        \item[\rm $\kappa$] covariance function
        \item[\rm $\theta$] parameters controlling a distribution family       \item[\rm $\lambda $] weight decay parameter
        \item[\rm $\mu$] mean
        \item[\rm $\sigma$] standard deviation
        \item[\rm $\phi(\cdot)$] activation function
        \end{description}
      \subsection*{Acronyms}
      \begin{description}
        \itemsep-0.2em\small
        \item[\rm AMI]      Adjusted Mutual Information
        \item[\rm ANN]      Artificial Neural Network
        \item[\rm AP]       Affinity Propagation
        \item[\rm ARI]      Adjusted Rand Index
        \item[\rm BNN]      Bayesian Neural Network
        \item[\rm CA]       Control Assembly
        \item[\rm CI]       Confidence Interval
        \item[\rm DNN]      Deep Neural Network
        \item[\rm FA]       Fuel Assembly
        \item[\rm GP]       Gaussian Process
        \item[\rm MCD]      Monte Carlo Dropout
        \item[\rm ML]       Machine Learning
        \item[\rm MTR]      Material Testing Reactor
        \item[\rm NN]       Neural Network
        \item[\rm NMI]      Normalized Mutual Information
        \item[\rm NRMSE]    Normalized Root Mean Squared Error
        \item[\rm PCA]      Principal Component Analysis 
        \item[\rm ReLU]     Rectified Linear Unit
        \item[\rm RI]       Rand Index
        \item[\rm SAFARI-1]	South African Fundamental Atomic Research Installation, generation 1
        \item[\rm UQ]       Uncertainty Quantification
        \item[\rm V\&V]     Verification and Validation
      \end{description}
    \end{multicols}
  \vspace{-2mm}
  \end{framed}
\end{table*}



\section{Introduction}
\label{section:introduction}

\subsection{The SAFARI-1 Research Reactor Measurement Dataset}

\begin{figure}[htb]
  \centering
  \raisebox{0.125\height}{\includegraphics[height=6.5cm]{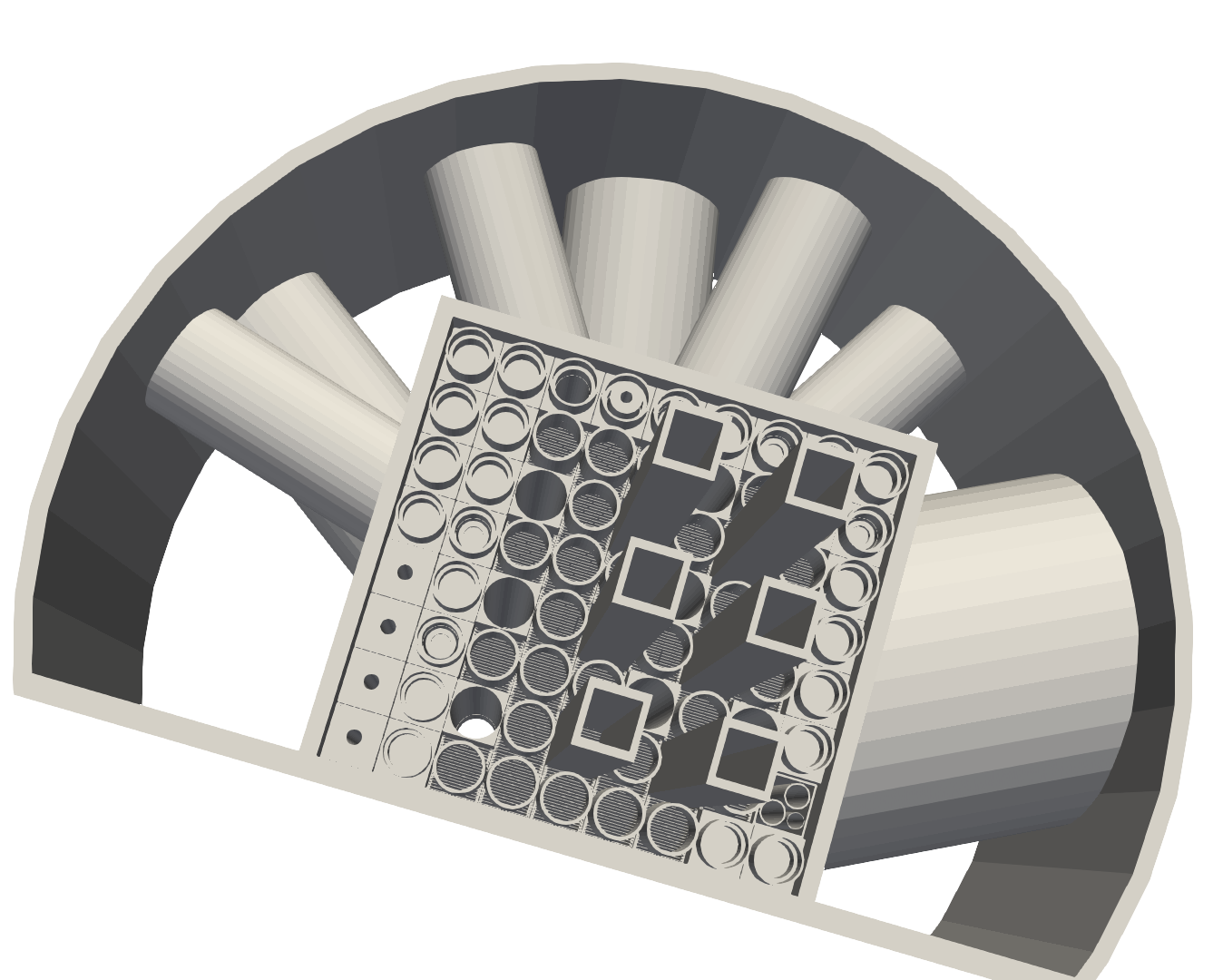}}
  \includegraphics[height=8cm]{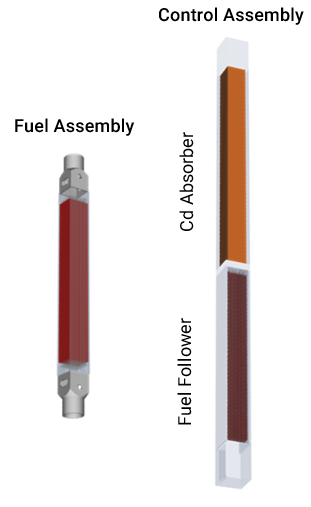}
  \caption{OSCAR-5 \cite{Prinsloo2017} models of the SAFARI-1 core (top view with partially extracted control rods), fuel and control assemblies.}
  \label{figure:The-SAFARI-1-core-and-assemblies}
\end{figure}

\begin{figure}[htb]
  \centering
  \includegraphics[clip, trim=0.1cm 0.2cm 0.1cm 0.1cm, width=0.6\linewidth]{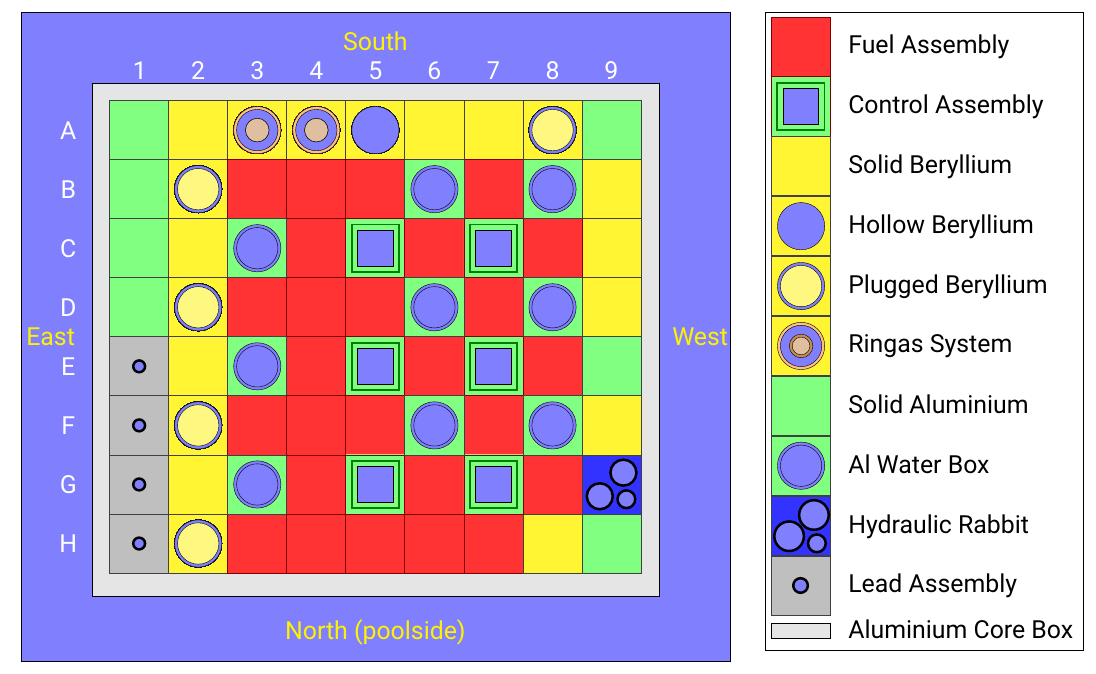}
   \caption{SAFARI-1 reactor core layout \cite{Moloko2023Prediction}.} 
  \label{figure:The-SAFARI-1-core-layout}
\end{figure}

The SAFARI-1 (South African Fundamental Atomic Research Installation, generation 1) research reactor is a 20 MW tank-in-pool, Material Testing Reactor (MTR) operated by the South African Nuclear Energy Cooperation SOC Ltd (Necsa). Its $9\times8$ core grid houses a total of 32 fuel-containing assemblies, of which 26 are MTR plate-type Fuel Assemblies (FAs) and six are follower-type Control Assemblies (CAs), also called control follower assemblies. Each follower-type CA consists of a top absorber and bottom fuel-follower section, connected by an aluminium coupling piece. The three-dimesional models of the SAFARI-1 reactor, as well as the FAs and CAs, are shown in Figure~\ref{figure:The-SAFARI-1-core-and-assemblies} and the two-dimensional core layout shown in Figure~\ref{figure:The-SAFARI-1-core-layout}.

\begin{figure}
  \centering
  \begin{subfigure}{0.495\linewidth}
    \centering
    \includegraphics[clip, trim=0.0cm 0.0cm 0.0cm 0.0cm, width=0.99\linewidth]{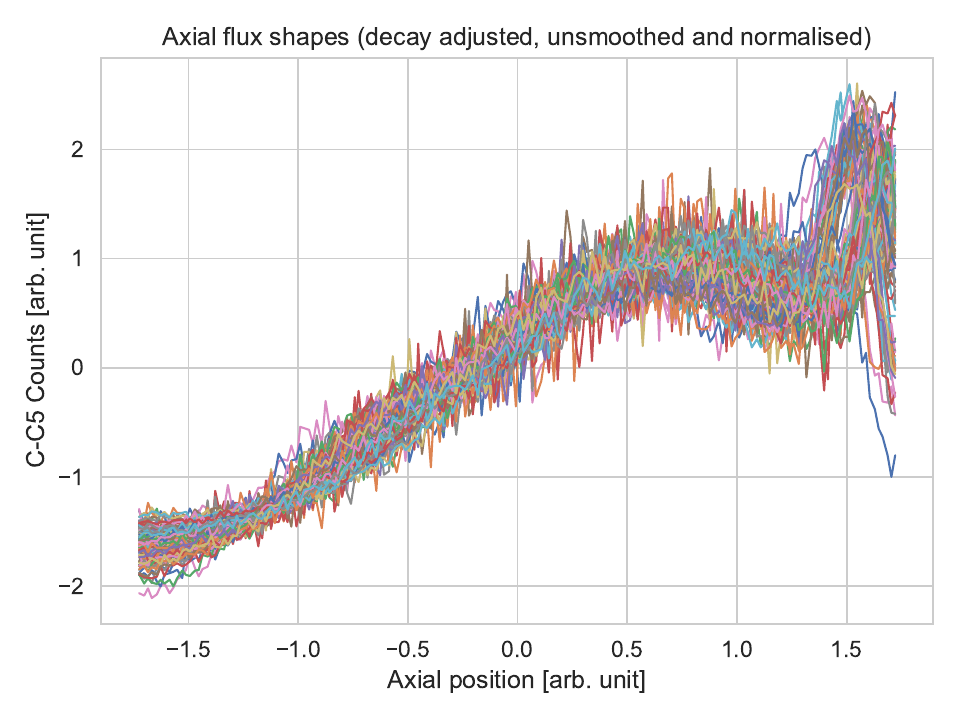}
    \caption{}
    \label{figure:flux-profiles-unsmoothed}
  \end{subfigure}
  \begin{subfigure}{0.495\linewidth}
    \centering
    \includegraphics[clip, trim=0.0cm 0.0cm 0.0cm 0.0cm, width=0.99\linewidth]{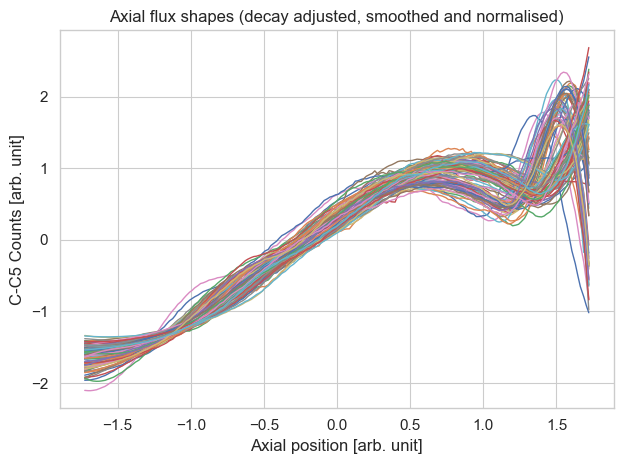}
    \caption{}
    \label{figure:flux-profiles-smoothed}
  \end{subfigure}
  \caption{Historical measured neutron flux profiles in the C5 control follower assembly before (a) and after (b) noise filtering. Both counts and axial positions are $z$-score normalized.
  }
  \label{figure:flux-profiles}
\end{figure}

Axial neutron thermal flux profiles are measured for each of the 32 fuel-containing assemblies at the beginning of each cycle by activation of the natural copper wires. The measured thermal flux profiles are utilized to estimate the reactor safety parameters, such as fuel peak-clad temperature, assembly power, etc. The measurement process is conducted manually by the reactor operators. The manual operation of the measurement process is associated with several challenges of a practical nature (duration of irradiation and cooling time, placement of a copper wire into the correct location within a fuel assembly, etc.), which may have an impact on the quality of the measured data and subsequently have an adverse effect on the estimation of the reactor safety parameters. Due to these challenges, the measured flux data contains statistical noise, measurement points may sometimes be missing, flux profiles may be shifted due to inadvertently displaced wires, etc. The ensemble of $z$-score normalized flux profiles, measured in the CA-housing core position C5 (see Figure~\ref{figure:The-SAFARI-1-core-layout}), over multiple historical cycles is shown in Figure~\ref{figure:flux-profiles-unsmoothed} for the sake of illustration.

\subsection{Issues Identified in Previous ML Research}

\begin{figure}
  \centering
   	\begin{subfigure}{1\linewidth}
		\centering
        \includegraphics[clip, trim=0.0cm 0.3cm 0.0cm 0.3cm, width=0.99\textwidth]{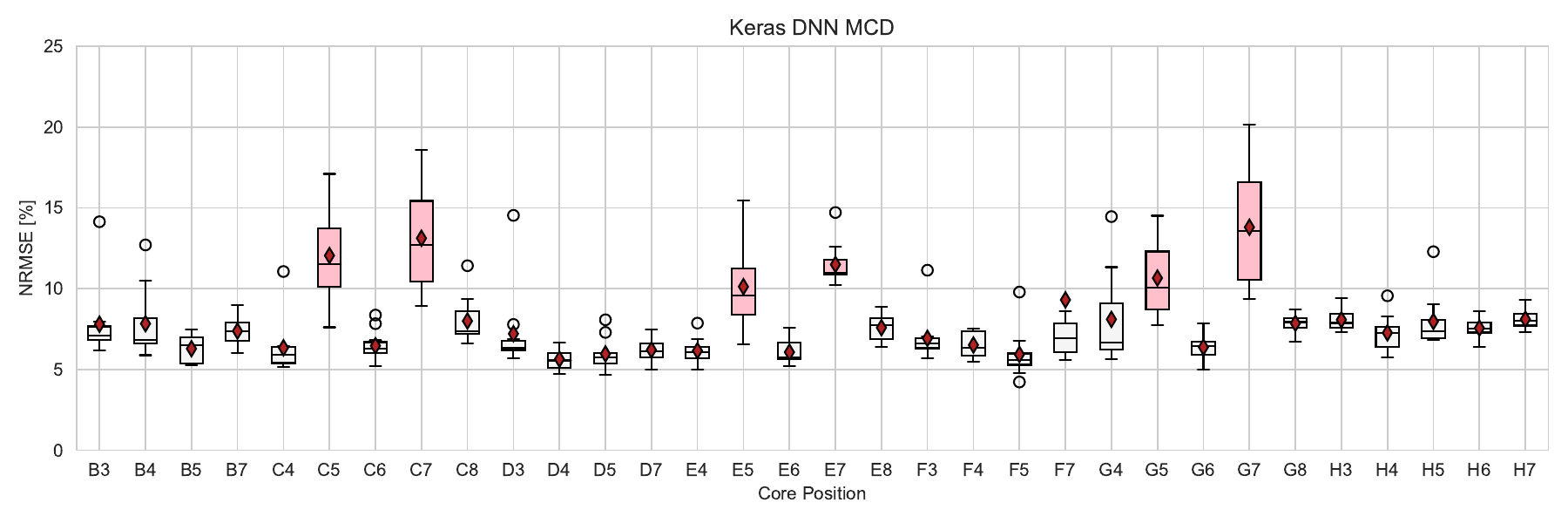}
        \caption{}
	\end{subfigure}
  	\begin{subfigure}{1\linewidth}
		\centering
        \includegraphics[clip, trim=0.0cm 0.3cm 0.0cm 0.3cm, width=0.99\textwidth]{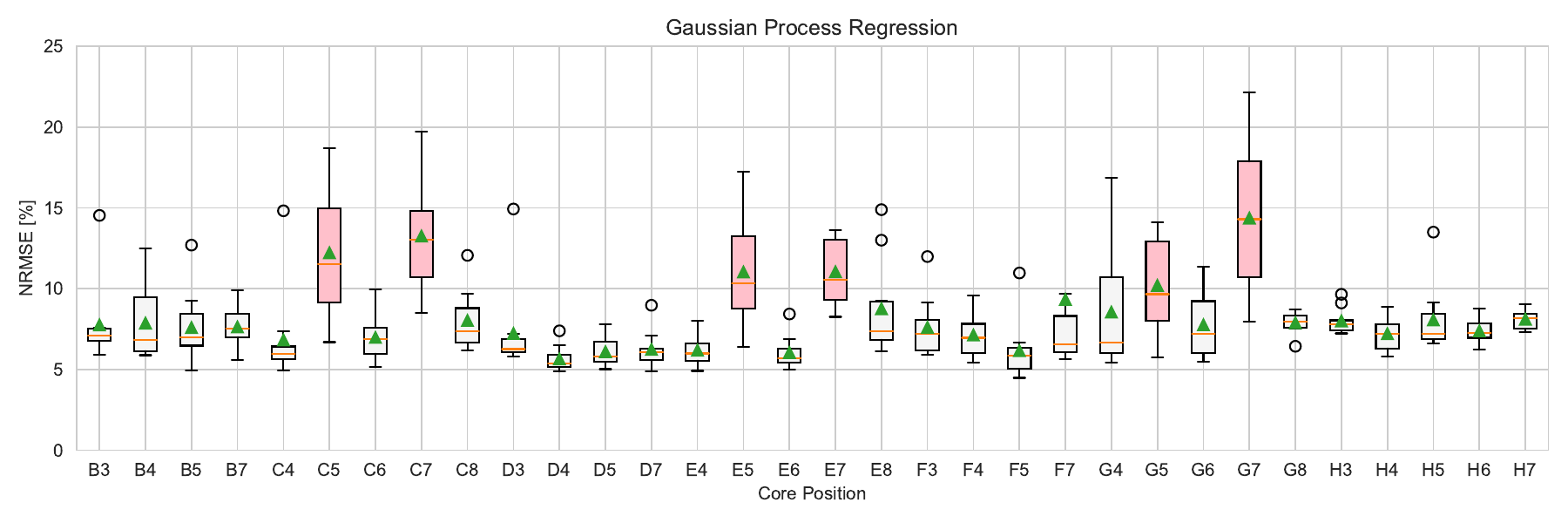}
		\caption{}
	\end{subfigure}
  \caption{Percentage NRMSEs of thermal flux shape prediction for FAs (white) and CAs (pink), using both DNN and GP ML models.}
  \label{figure:Percentage-Normalized-Root}
\end{figure}

Artificial Neural Network (ANN) models, trained on the copper-wire measurement data, have been developed in our previous work \cite{Moloko2023Prediction,Moloko2021Estimation,Moloko2022Quantification} to predict the axial flux shapes. This additional information would help to detect the deficiencies in measured data and eventually to correct them \cite{Moloko2021Estimation}. The data-driven Machine Learning (ML) models could also be useful in the validation of neutronic codes by providing an indication of the quality of a specific experimental data set. The ANN models have generally yielded predictions that agree well with the measured axial neutron flux profile data, especially in the 26 FAs. However, it was observed that the ANN model predictions for the six CAs resulted in large discrepancies of up to \SI{23}{\percent} (in terms of the Normalized Root Mean Squared Error -- NRMSE), when compared to the measurement data. As a comparison, the prediction errors obtained for the FAs were typically between \SI{5}{\percent} and \SI{10}{\percent}. This can be clearly seen in Figure~\ref{figure:Percentage-Normalized-Root} where the distribution of NRMSE over ten operational cycles (which constitute a prediction dataset not used in training) is represented by box plots for all the fuel-containing assemblies. In this figure, a Tukey-style box plot convention is employed with the only non-standard element -- the mean -- presented with diamond or triangle markers, respectively.

In Figure~\ref{figure:Percentage-Normalized-Root}, the top graph is a result from our the previous studies \cite{Moloko2023Prediction,Moloko2022Quantification} and corresponds to a model obtained for the Deep Neural Network (DNN) with Monte Carlo Dropout (MCD), whereas the bottom graph is for the Gaussian Process (GP) produced as part of this study (more detail on the GP will be provided in Section~\ref{section:methodologies-GP}). As may be noticed, the results from DNN and GP are overall consistent for most of the assemblies with some minor differences. This leads to the conclusion that the lower accuracy of prediction for CAs is not an artefact of the ML algorithm (along with selected hyperparameters) being used. Neither of this can be explained by the statistical noise (the major contributor to the discrepancy between ML model predictions and measurements for FAs), since the signal-to-noise ratio is smaller for CAs than for FAs.

\subsection{Analysis of Identified Issues, Proposed Solution and Objective of this Study}

Our recent analyses \cite{Moloko2023Improving} revealed that the reduced ML model prediction accuracy of the CAs can be attributed to the presence of at least two distinct, yet unexplained, groups (hereinafter referred to as \emph{clusters}) of axial profile shapes in the CAs' measured data. These clusters of axial profiles can be clearly distinguished after applying noise filtering, as is shown in Figure~\ref{figure:flux-profiles-smoothed} for the dataset from Figure~\ref{figure:flux-profiles-unsmoothed}. In Figure~\ref{figure:flux-profiles-smoothed}, one may see a group of profiles with a peak near the right boundary of the figure, whereas another group does not have such a peak, and no continuous transition between these two groups can be observed. It is worth noting, that since the copper-wire measurements are performed in the bottom fuel-follower section of control rods (as illustrated in Figure~\ref{figure:The-SAFARI-1-core-and-assemblies}), this peak in the measured profile corresponds to the thermal flux peak in the coupling piece between the follower and the top absorbing section of CA.  

Training ML models for the CAs, based on these two clusters grouped together, makes it challenging to reach a high accuracy. This is because it is difficult to fit a ML model that simultaneously agrees well with two clusters of data that have very different shapes. 
Therefore, to improve the prediction of axial flux profiles in the CAs, we propose to develop enhanced ML models by combining supervised and unsupervised ML. With unsupervised ML, clustering algorithms will be first employed to identify clusters in the measured axial neutron flux profiles. Then, regression-based supervised ML models, including DNN and GP, will be trained for each of the identified clusters separately. To improve the trustworthiness of the ML models' predictions, their approximation uncertainties will also be quantified and compared.

For clustering analysis, we have selected the $k$-means \cite{Macqueen1967Some,Hartigan1979Algorithm} algorithm due to its simplicity, and the AP \cite{Frey2007Clustering} algorithm due to its advantage of not requiring the user to specify the number of clusters before training. Both approaches will be used to identify subsets of activation shapes. 
The theory behind these two clustering techniques will be briefly discussed in Section~\ref{section:methodologies}, and we finalize this section with a brief literature survey with a special focus on the application of the clustering methods and Uncertainty Quantification (UQ) of ML models in nuclear engineering-related studies.

\subsection{Literature Survey}



The $k$-means and AP clustering algorithms have been widely used in various research fields, often in combination with or as a component of other algorithms, and their effectiveness, drawbacks, and solutions have been reported \citep{Zhu2009kmeansAP,Chen2014AP-PCA,Ahmad2019SurveyClustering,Ahmed2020kmeans,Duan2023AffinityPropagation}.
%
%
In \cite{Chen2014AP-PCA}, a combination of Principal Component Analysis (PCA) and AP clustering is proposed to reduce the misclassification of points in high-dimensional data. PCA is used to reduce the dimensionality of the original data while still preserving the majority of the information of the variables. Subsequently, AP clustering is applied in low-dimensional space. Experiments have shown that this method is effective in improving classification accuracy.
%
%
The authors of \cite{Liu2020AffinityPropagation} propose a novel dynamic optimization method based on AP clustering to track moving optima and a self-adaptive strategy based particle swarm optimizer to quickly find the peaks in each subspace. 
%
%
In \cite{Song2022DT}, an autonomous calibration method for the digital twin of nuclear power plants is proposed to compensate for the error in the results of the low-accuracy digital twin. ANN models are trained with clustered samples generated using $k$-means clustering, and real data. The method has been used successfully on measured data from the steam generator pipe rupture experiment. The work presented in \cite{Anupreethi2020kmeans} proposes a data-driven approach based on in-core detector measurements to optimize the number and locations of in-core detectors in an Advanced Heavy Water Reactor. The $k$-means clustering algorithm is used to identify linear relationships among the detectors and group them into clusters. 

UQ is the process to estimate uncertainties in the model outputs by propagating uncertainties from input parameters. UQ of physics-based models have been widely studied in nuclear engineering over the last three decades. However, UQ of data-driven ML models is important but less studied, particularly in nuclear engineering. ML-based models are subject to approximation uncertainties due to data noise, incomplete domain coverage, and imperfect models, which are expected to be larger when training data are limited and predictions are made in extrapolated regions. The uncertainties associated with ML models can be divided into two categories: aleatoric and epistemic. Aleatoric uncertainty is caused by noisy data and can be categorized into homoscedastic and heteroscedastic uncertainty. Epistemic uncertainty is the uncertainty associated with the model and can be reduced with more data. Both aleatoric and epistemic uncertainty should be used to estimate predictive uncertainty and quantify confidence in model predictions \citep{Abdar2021UQDeepLearning,Psaros2023UQ,Yaseen2023VVUQ}. 


Recent advances in UQ methods used in deep learning have been applied to a variety of real-world problems in science and engineering, such as computer vision, image processing, medical image analysis, natural language processing, and bioinformatics. These advances are reviewed in \cite{Abdar2021UQDeepLearning}.
Dropout, a technique used to avoid overfitting, is used as an approximate Bayesian Inference in \cite{gal2016dropout} and ``Bayes by Backprop'' (i.e. uncertainty on the weights of a neural network) is utilized in \cite{blundell2015weight} to represent model uncertainty. Both studies show that UQ is essential for both classification and regression problems.

In nuclear engineering-related studies, \cite{Yaseen2023VVUQ} explores and compares three different techniques for UQ of DNNs: MCD, deep ensembles (DEs) \citep{lakshminarayanan2016UQ}, and Bayesian neural networks (BNNs). These methods are combined with PCA for time-dependent simulations, using the fission gas release model in the BISON code. The paper demonstrates the feasibility of combining supervised ML (regression using DNN) with unsupervised ML (dimensionality reduction using PCA) for UQ of neural network models. 
In our previous work \cite{Moloko2022Quantification}, application of DNN to predict assembly axial neutron flux profiles in the SAFARI-1 research reactor and quantification of prediction uncertainties using MCD and BNN solved by Variational Inference (BNN VI) was successfully demonstrated. 
%
%
Furthermore, \cite{Moloko2023Improving} used the same data with $k$-means and AP, to identify clusters in the measured axial neutron flux profiles and train on the identified clusters. The prediction uncertainties based on MCD and GP models is also provided.

Despite the plurality of UQ methods, there are several open challenges and gaps in the field of UQ, such as dealing with large measurement noise levels, selecting and tuning hyperparameters, obtaining representative samples from multimodal posterior distributions, evaluating performance, predicting correlations between inputs, and assessing robustness in cases of distribution shift. Some of these challenges are discussed in \cite{Abdar2021UQDeepLearning,Psaros2023UQ}. 
%
UQ provides a tangible, numerical way to measure the accuracy of the model's predictions. This paper employs two methods, previously used in \cite{Moloko2023Improving}, namely MCD and GP, to quantify ML models prediction uncertainties.

The rest of this paper is organized as follows. In Section~\ref{section:data-processing}, historical copper-wire measurement data, used in this study, are presented and their pre-processing discussed. In Section \ref{section:methodologies}, we briefly describe the ML algorithms employed in this work, including clustering analysis and regression techniques, as well as the UQ approaches. Results of applying the proposed combination of supervised and unsupervised ML are presented and discussed in Section~\ref{section:results-and-discussion}. Section~\ref{section:conclusions} summarizes our major findings and concludes this paper.

\section{Pre-processing of Data}
\label{section:data-processing}

In this work, the pre-processed training data from previous studies \cite{Moloko2023Prediction,Moloko2022Quantification} was utilized. Using the same training data allows a direct comparison of new and old flux shape prediction results. To make this manuscript self-contained, we provide  a brief description of the data and its pre-processing in this section. 

As in our previous study \cite{Moloko2022Quantification}, the copper-wire measurement data from 86 cycles were subdivided into 76 cycles selected for training, validation and testing, whereas ten cycles were reserved for evaluating the generalization and prediction capability of the trained ML models. For each operational cycle and for each assembly, the detector counts are measured at 180 equidistant axial locations along the active height and 86 cycles were selected for the study; at most 13\,680 data points are available for training for a specific assembly. Here we use specifier ``at most'' because measurements for a given assembly and cycle may contain less than 180 axial data points. Herein, a data point is defined as a tuple containing three numbers: 1) control bank position, 2) axial location of measurement point, and 3) detector count value. When training the ML models, the first two numbers are treated as input and the third number is the output. 

The data pre-processing procedure included exponential decay-corrections and $z$-score normalization (standardization) of the input and output data as is recommended in the literature \cite{Nawi2013TheEffect} and done in our previous studies \cite{Moloko2023Prediction,Moloko2021Estimation,Moloko2022Quantification}. The decay correction was applied for all the wires of each cycle and consisted in decaying all counts back to the time of the beginning of scanning of the first wire. The normalization was performed for each of 32 measured profiles and each cycle separately.  

To be able to use the incomplete measurements for clustering analysis, the missing points in flux profiles were filled with linearly interpolated or extrapolated values. This data completion procedure was preceded by smoothing the pre-processed data with the Savitzky-Golay filter as described in \cite{Moloko2023Prediction,Moloko2021Estimation}. It is worth noting that our attempts to use the raw noisy measured data yielded poor results in terms of cluster identification. \emph{Smoothing} and \emph{filling the missing data} may therefore be regarded as two additional data pre-processing steps, required for the clustering analysis reported in this work.

\section{Methodologies}
\label{section:methodologies}

\subsection{Overview}

ML models typically make predictions on data using computer algorithms that improve automatically through experience (data). When the dataset consists of labels along with data points, i.e. input-output pairs, it is known as \emph{supervised learning}. ML models analyse the training data to learn function mappings from the measurable input properties or characteristics, referred to as \emph{features}, to the target output. Examples are \emph{regression and classification}. On the other hand, the ML methods, where insights are drawn from data points without any ground truth or correct labels, fall under the category of \emph{unsupervised learning}, which is widely used in exploratory data analysis to make sense of the data before training more complex ML models to make inferences. Examples of unsupervised ML are \emph{dimensionality reduction} and \emph{clustering}. In these processes, the dataset is either unlabelled or labelled but one does not seek to learn the input-output mapping in the dataset.

Our previous work \cite{Moloko2023Prediction,Moloko2021Estimation,Moloko2022Quantification} focused on developing DNN-based regression models to learn the input-output mapping in the dataset only, which falls within the realm of supervised ML. In this work, we develop a procedure that combines unsupervised and supervised ML to address the issues noticed in the CAs. DNN and GP-based regression models are employed for supervised learning, while $k$-means- and Affinity Propagation clustering algorithms are used for unsupervised learning. Given that the problem under consideration does not have a reference solution, we selected two ML algorithms for both tasks to ensure that we can obtain consistent results using different models. In Section~\ref{section:methodologies-clustering}, we will provide an overview of clustering, as well as the $k$-means and Affinity Propagation (AP) algorithms. Section~\ref{section:methodologies-GP} includes a brief introduction of the GP theory, how data uncertainty is included in the regression process and how the GP prediction uncertainty is quantified. Finally, Section~\ref{section:methodologies-DNN-MCD} presents a concise discussion of DNN-based regression and the adoption of MCD for UQ of DNN predictions.

\subsection{Clustering Analysis}
\label{section:methodologies-clustering}

\subsubsection{Overview of clustering}

Clustering analysis, also called data segmentation, is a technique used in data mining and ML to collect similar objects into groups, also called \emph{clusters} \cite{Hastie2009Elements}. A cluster refers to a collection of data points aggregated together because of certain similarities. The objective of clustering is to divide the dataset into several groups in such a way that the data points in the same groups are more similar to each other than to the data points in other groups. The training process (the process to assign the training samples to appropriate clusters) usually involves maximizing inter-cluster distances and minimizing intra-cluster distances. 

Clustering algorithms can be generally divided into three categories:
\begin{itemize}
    \item[(1)] \emph{partitional clustering}, that divides data objects into non-overlapping groups through an iterative process to assign subsets of data points into $k$ clusters, such as $k$-means and $k$-medoids, 
    \item[(2)] \emph{hierarchical clustering}, that determines cluster assignments by building a hierarchy, such as agglomerative clustering (bottom-up) and divisive clustering (top-down), and 
    \item[(3)] \emph{density-based clustering}, which determines cluster assignments based on the density of data points in a region, such as Gaussian mixture models.
\end{itemize}
Note that this categorization is not unique; for instance, a significantly more extensive list of categories is discussed in \cite{Xu2015AComprehensive}. In terms of the above categorization, both $k$-means and AP, selected for this study, belong to the partitional clustering. 

\subsubsection{Clustering with $k$-means}

The $k$-means is one of the most popular clustering algorithms because of its simplicity. It uses $k$ centroids to define clusters, which are the imaginary or real locations (which depends on whether a centroid is a ``real'' sample point in the dataset) representing the centre of the cluster. A data point is considered to be in a particular cluster if it is closer to that cluster's centroid than any other centroids. $k$-means uses an iterative process to assign each data point to the groups and slowly data points get clustered based on similar features. The correct group, each data point should belong to, is identified by minimizing the sum of distances (e.g. Euclidean distance, Manhattan distance) between the data points and the cluster centroid. Advantages of this algorithm include: the simplicity of understanding and implementation, computational efficiency and scalability to large data sets, guaranteed convergence; whereas drawbacks include the manual selection of the number of clusters, lack of consistency -- dependence on the initial centroid selection, sensitivity to outliers and to scale.

\subsubsection{Clustering with AP}

In contrast to traditional clustering methods, AP does not require one to specify the number of clusters. It is based on the concept of ``message-passing'' between the data points \cite{thavikulwat2008affinity}. AP creates clusters by sending messages between data points until convergence. A dataset is described using a small number of \emph{exemplars}, which are members of the dataset that are representative of clusters. With AP, each data point sends messages to all other points informing its targets of each target's relative attractiveness to the sender. Each target then responds to all senders with a reply informing each sender of its availability to associate with the sender, given the attractiveness of the messages that it has received from all other senders. Senders reply to the targets with messages informing each target of the target's revised relative attractiveness to the sender, given the availability messages it has received from all targets. The message-passing procedure proceeds until a consensus is reached. Once the sender is associated with one of its targets, that target becomes the point's exemplar. All points with the same exemplar are placed in the same cluster.

AP clustering does not need to set the number of clusters, and has advantages on efficiency and accuracy. It has, however, some deficiencies \cite{Xu2015AComprehensive}: it can be computationally expensive, especially for large datasets, making it unsuitable for large-scale clustering problems. It can be sensitive to the choice of a metric used as the measure of similarity between data points. In addition, though the number of clusters does not need to be pre-set by the user, the ``sample preference'' and ``damping'' hyperparameters still need to be specified. The ``sample preference'' controls how many exemplars are used, while the ``damping factor'' damps the responsibility and availability of messages to avoid numerical oscillations when updating these messages. Nevertheless, when oscillations occur, AP cannot automatically eliminate them.

\subsubsection{Hyperparameters and metrics}

In this study, the Euclidean distance between points in a 180-dimensional space (this corresponds to the number of data points in a flux shape) was used as the similarity measure for both clustering algorithms. Hyperparameters of both clustering algorithms (namely, the number of clusters $k$ in $k$-means, and the sample preference and damping factor for AP) have been tuned until they provided consistent results, which were checked by a visual inspection and by evaluating three similarity scores, namely, the Adjusted Rand Index (ARI) \cite{Rand1971Objective,Kuncheva2004UsingDI}, the Adjusted Mutual Information (AMI) and the Normalized Mutual Information (NMI) \cite{Strehl2002Cluster} indicators \cite{JMLR:v12:pedregosa11a}. Each indicator computes a similarity measure between two clusterings by considering all pairs of samples and counting pairs that are attributed to the same or different clusters. By the construction, they have a value close to 0.0, for random labelling independently of the number of clusters and samples, and exactly 1.0 when the clusters are identical up to a permutation. Definitions of the three employed similarity indices are provided in \ref{sec:Appendix-A}.

Note that there are other widely used clustering validity measures \cite{JMLR:v12:pedregosa11a}, such as the Dunn index, the Davies-Bouldin index, the Silhoutte index, the Calinski-Harabasz index, etc. In this work, we have selected ARI, AMI and NMI because they have clear definitions in the range of $[0,1]$, making them easier for a consistency check between $k$-means and AP.

\subsection{Gaussian Process Regression}
\label{section:methodologies-GP}

\emph{Gaussian Process} (GP) modeling, also known as \emph{Kriging}, or \emph{spatial correlation modeling}, was originally developed by geologists in the 1950s to predict the distribution of minerals over an area of interest given a set of sampled sites. It was made popular in the context of modeling \& simulation and optimization by Sacks et al.~\cite{sacks1989design}  and Jones et al.~\cite{jones1998efficient}, respectively. GP has been widely used as a ML method \cite{Rasmussen2005Gaussian} to construct metamodels for computer models in many areas. A GP model is a generalized linear regression model that accounts for the correlation in the residuals between the regression model and the observations. The only assumption for GP modeling is that the model response is continuous and smooth over the domain, which is true for most problems. Thus, if two points are close to each other in the input domain, the residuals in the regression model should be close. It follows that we do not treat the residuals as independent by assuming that the correlations between the residuals are related to the distance between the corresponding inputs.

A GP is fully specified by a \emph{mean function} $m(\bm{x})$ and a positive definite \emph{kernel function} or \emph{covariance function} $\kappa(\bm{x},\bm{x}')$ that sets the covariance between two points $\bm{x}$ and $\bm{x}'$ \cite{Krasser2018Gaussian}:
\begin{equation}
  f(\bm{x}) = \mathcal{GP}\left(m(\bm{x}), \kappa(\bm{x},\bm{x}')\right), \quad \bm{x} \in \mathbb{R}^n.
  \label{eq:joint-pdf-Normal}
\end{equation}
By this definition, any collection of function values $\{f(\bm{x}_1),\ldots,f(\bm{x}_N)\}$, indexed by their corresponding inputs $\bm{X} = \{\bm{x}_1,\ldots,\bm{x}_N\}$, have a joint multivariate Normal distribution:
\begin{equation}
  p(\bm{f} \lvert \bm{X}) = \mathcal{N}(\bm{f} \lvert \boldsymbol\mu, \bm{K}),
  \label{eq:GP-prior}
\end{equation}
where $\bm{f} = (f(\bm{x}_1),\ldots,f(\bm{x}_N))$, $\bm\mu = (m(\bm{x}_1),\ldots,m(\bm{x}_N))$ and $K_{ij} = \kappa(\bm{x}_i,\bm{x}_j)$. The mean function is typically taken to be zero, since GPs are sufficiently flexible to model the mean arbitrarily well, hence we will assume hereinafter that $\bm\mu = \bm{0}$. Equation~\eqref{eq:GP-prior} specifies a GP prior.
Given a training dataset with function values $\bm{f}$ at inputs $\bm{X}$, a GP prior can be converted into a GP posterior $p(\bm{f}_* \lvert \bm{X}_*,\bm{X},\bm{f})$ which can then be used to make predictions $\bm{f}_*$ at new inputs $\bm{X}_*$, a procedure referred to as \emph{GP regression}. By definition of a GP, the joint distribution of observed values $\bm{f}$ and predictions $\bm{f}_*$ is a Gaussian which can be partitioned as follows:
\begin{equation}
  \begin{pmatrix}
    \bm{f} \\ \bm{f}_*\end{pmatrix} \sim \mathcal{N}
  \left(\boldsymbol{0},
  \begin{pmatrix}\bm{K} & \bm{K}_* \\ \bm{K}_*^\intercal & \bm{K}_{**}\end{pmatrix}
  \right), \label{eq:Gaussian-partitioning}
\end{equation}
where $\bm{K}_* = \kappa(\bm{X},\bm{X}_*)$ and $\bm{K}_{**} = \kappa(\bm{X}_*,\bm{X}_*)$. With $N$ training data points and $N_{*}$ new input data points, $\bm{K}$, $\bm{K}_{*}$
and $\bm{K}_{**}$ are $N\times N$, $N\times N_{*}$ and $N_{*}\times N_{*}$
covariance matrices, respectively.
For a training dataset with noisy function values $\bm{y} = \bm{f} + \bm\epsilon$, where $\bm\epsilon \sim \mathcal{N}(\bm{0}, \sigma_y^2 \bm{I})$, noise is independently added to each observation ($\bm{I}$ is the $N \times N$ identity matrix).
Applying rules for conditioning Gaussians, the predictive distribution is given by
\begin{equation}
    \begin{aligned}
        p(\bm{f}_* \lvert \bm{X}_*,\bm{X},\bm{y}) &= \mathcal{N}(\bm{f}_* \lvert \bm{\mu}_*, \bm{\Sigma}_*),  \\
        \bm{\mu_*}  &=  \bm{K}_*^\intercal (\bm{K} + \sigma_y^2\bm{I})^{-1} \bm{y},  \\
        \bm{\Sigma_*} &= \bm{K}_{**} - \bm{K}_*^\intercal (\bm{K} + \sigma_y^2\bm{I})^{-1} \bm{K}_*.
    \end{aligned}
  \label{eq:conditional-mean-covariance}
\end{equation}

As can be seen, the prediction uncertainty of the GP model is directly available, making GP the only ML model with such a capability. This is the primary reason to use GP models to provide a benchmark comparison with the UQ results of the DNN models.
Although Equations~\eqref{eq:conditional-mean-covariance} cover noise in training data, it is still a distribution over noise-free predictions $\bm{f}_*$. To additionally include noise $\bm\epsilon$ into predictions $\bm{y}_*$, $\sigma_y^2$ needs to be added to the diagonal of $\bm{\Sigma_*}$:
\begin{equation}
  p(\bm{y}_* \lvert \bm{X}_*,\bm{X},\bm{y}) = \mathcal{N}(\bm{y}_* \lvert \bm{\mu}_*, \bm{\Sigma}_* + \sigma_y^2\bm{I}),
  \label{eq:9}
\end{equation}
using the definitions of $\bm{\mu}_*$ and $\bm{\Sigma}_*$ from \eqref{eq:conditional-mean-covariance}. For a more detailed theory of GP, interested readers can refer to Chapter 2 of \cite{Rasmussen2005Gaussian}. 

In this work, GP-based regression models were implemented in Python (Numpy library) with some amendments, discussed below. We used the squared exponential kernel also known as Gaussian kernel or Radial Basis Function kernel and given by
\begin{equation}
  \kappa(\bm{x}_i,\bm{x}_j) = \sigma_f^2 \exp\left(-\frac{(\bm{x}_i - \bm{x}_j)^\intercal
    (\bm{x}_i - \bm{x}_j)}{2l^2}
  \right) + \sigma_y^2\delta(\bm{x}_i,\bm{x}_j),
  \label{eq:rbf}
\end{equation}
where $\delta(\bm{x}_i,\bm{x}_j)$ is the Kronecker delta.  The length-scale parameter $l$, also called the characteristic length-scale, controls how far a sample point's influence extends: larger $l$ values result in higher correlation, indicating that output values are similar when two inputs are close, while smaller $l$ values lead to lower correlation, meaning that there will be large differences between outputs, even for nearby points. Also, the higher $l$ values, the narrower the uncertainty regions between training data points, while lower values result in wider uncertainty regions.
The hyperparameter $\sigma_f$ controls the vertical variation of functions drawn from the GP and $\sigma_y$ represents the noise in the training data. The higher values of $\sigma_y$ result in more coarse approximations which avoids over-fitting to noisy data. It is therefore important that this hyperparameters are chosen sensibly, i.e. optimized.

In this study, hyperparameters $l$ and $\sigma_f$ were adjusted manually, until satisfactory results in terms of the predicted flux shape have been achieved (note that we used the same value of the scale parameter $l$ for all input features). The parameter $\sigma_y$ was attributed point-wise as a square root of decay-corrected (see details in Section~\ref{section:data-processing}) count values. 

Data from 76 historical cycles were employed for training the ML models (more precisely, training, validation and testing, as discussed in Section~\ref{section:data-processing}), and the remaining ten cycles were reserved for evaluating the ML models' prediction capability.

\subsection{Deep Neural Network with Monte Carlo Dropout}
\label{section:methodologies-DNN-MCD}

An ANN is an information processing model, which consists of a large number of interconnected units, referred to as \emph{neurons} or \emph{nodes}, organized in layers, that is, input layer, output layer, and one or more hidden layer(s) between the input and output layers \cite{Suzuki2011Artificial}.
An ANN with multiple hidden layers is also referred to as a DNN. DNNs differ from traditional ML techniques, in that they can automatically learn representation from data (automatic feature extraction) without introducing hand-coded rules or human domain knowledge \cite{Lecun2015Deeplearning}. The two common issues in training DNNs are the potential vanishing gradient and over-fitting. The vanishing gradient problem can been addressed by using a good set of activation functions, e.g.~the ReLU activation function, and the over-fitting issue can be alleviated using $L_1$, $L_2$ as well as the dropout regularization techniques, thus allowing the model to perform better \cite{Lecun2015Deeplearning,Saptarshi2020Deeplearning,Srivastava2014dropout,Liu2018DNN}.

The standard DNN only predicts deterministic outputs given an input, without providing the model prediction uncertainties. This is because the DNN training process only provides deterministic estimates of the DNN parameters (weights and biases). In this study, the MCD technique \cite{gal2016dropout}  is employed to quantify uncertainties of the DNN predictions. 
In MCD, dropout is used to introduce randomness to both the training and prediction processes. Once trained, the DNN can be evaluated for the same input multiple times while dropping the hidden neurons randomly, resulting in a collection of predictions that can be used to estimate mean values and variances as an indication of uncertainties.

An MCD process is obtained by adding dropout layers and introducing scaling factors between layers in a standard DNN. In MCD, for a DNN of depth $L$, with a dropout probability $p$, the loss function is given by:
\begin{equation}
   \begin{aligned}
           \mathcal{L}_{\text{dropout}}
           = \frac{1}{N}\sum_{i=1}^{N}
           {\left\Vert \hat{\bm{y}}_{i}-\bm{y}_{i}\right\Vert ^{2}}
           +\lambda\sum_{\ell=1}^{L}\left(p\left\Vert \bm{W}_{\ell}\right\Vert _{2}^{2}+\left\Vert \bm{b}_{\ell}\right\Vert _{2}^{2}\right),
     \end{aligned}\label{equation:mcdloss}
\end{equation}
where $\bm{W}_\ell$ is the matrix of weights and $\bm{b}_\ell$ is the vector of biases at hidden or output layer $\ell=1,\ldots,L$, $\hat{\bm{y}}$ and $\bm{y}$ are the prediction and target output of the DNN for input $\bm{x}$, respectively, and $\lambda$ is the weight decay coefficient. In MCD, the mean prediction ($\hat{\bm{\mu}}$) is estimated by averaging over $T$ forward passes or runs while the uncertainty can be estimated in terms of the standard deviation  ($\hat{\sigma}$):

\begin{equation}
    \begin{aligned}
        \hat{\bm{\mu}}(\bm{x}) =\frac{1}{T}\sum_{t=1}^{T}\hat{\bm{y}}^{(t)}(\bm{x}),\quad
        \hat{\sigma}^{2}(\bm{x}) = \frac{1}{T-1}\sum_{t=1}^{T}\left(\hat{\bm{y}}^{(t)}(\bm{x})-\hat{\bm{\mu}}(\bm{x})\right)^{2}.
    \end{aligned}
\end{equation}

DNN regression models made use of data points from 76 training cycles, randomly partitioned into \SI{80.75}{\percent} for training, \SI{5}{\percent} for testing and \SI{14.25}{\percent} for validation. A DNN model was constructed for each of the six CAs and trained for each of the identified clusters. Each DNN model consists of an input layer with two neurons, corresponding to the control bank position and the axial location along the wire, one output layer with a single neuron representing a count value (an indicator for the axial neutron flux value) and several hidden layers. For each of the six CA models, hyperparameters were obtained and the Keras TensorFlow Probability library was used to build the DNN models with the capability to perform UQ of the DNN predictions. DNN training was performed using the Adam optimizer and ReLU activation function in the hidden layers. The prediction mean values were estimated by averaging over \num{20000} DNN runs and the uncertainties were estimated in terms of the standard deviations. The analyses and results of this case are conducted using Python under the Keras deep learning package with the TensorFlow backend \cite{chollet2015keras}.

\section{Results and Discussions}
\label{section:results-and-discussion}

Results of this study are divided into two parts: 1) clustering analysis conducted using the $k$-means and AP methods in Section~\ref{section:results-and-discussion-clustering}, and 2) axial flux profile predictions and the associated uncertainties, using the DNN (with MCD) and GP models in Section~\ref{section:results-and-discussion-regression-and-UQ}. In order to demonstrate the improvement in prediction accuracy, the results using measurement data from the CAs before and after clustering analysis will be compared. The DNN and GP predictions are based on the ten SAFARI-1 operation cycles, which were not used in training/validation/testing of the ML models. Due to the large volume of generated results (six CAs for ten cycles), the axial flux profiles and UQ results of assemblies C5 and E5 for one selected cycle (i.e. cycle C2010-1) will be presented as a representative case. Results from other CAs and cycles have shown similar behaviour. 

\subsection{Results for Clustering Analysis}
\label{section:results-and-discussion-clustering}

\begin{figure}[!htb]
	\centering
    \begin{subfigure}{0.575\textwidth}
        \centering
        \includegraphics[width=1\linewidth]{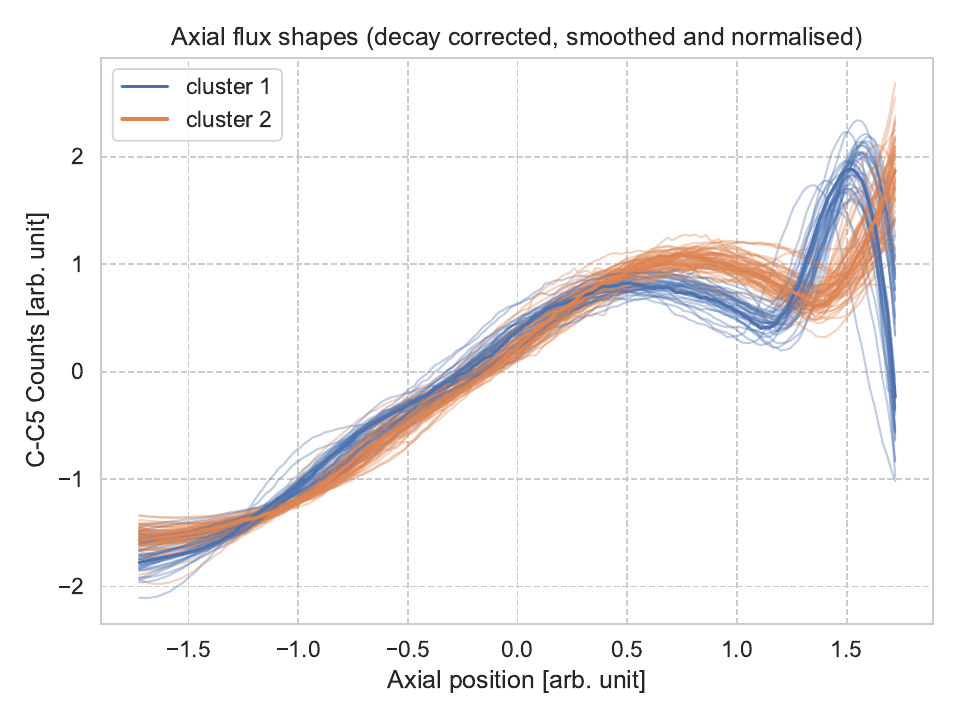}
        \caption{}\label{figure:cluster-partitioning-C5-a}
    \end{subfigure}
    \begin{subfigure}{0.41\textwidth}
        \centering
        \includegraphics[width=1\linewidth]{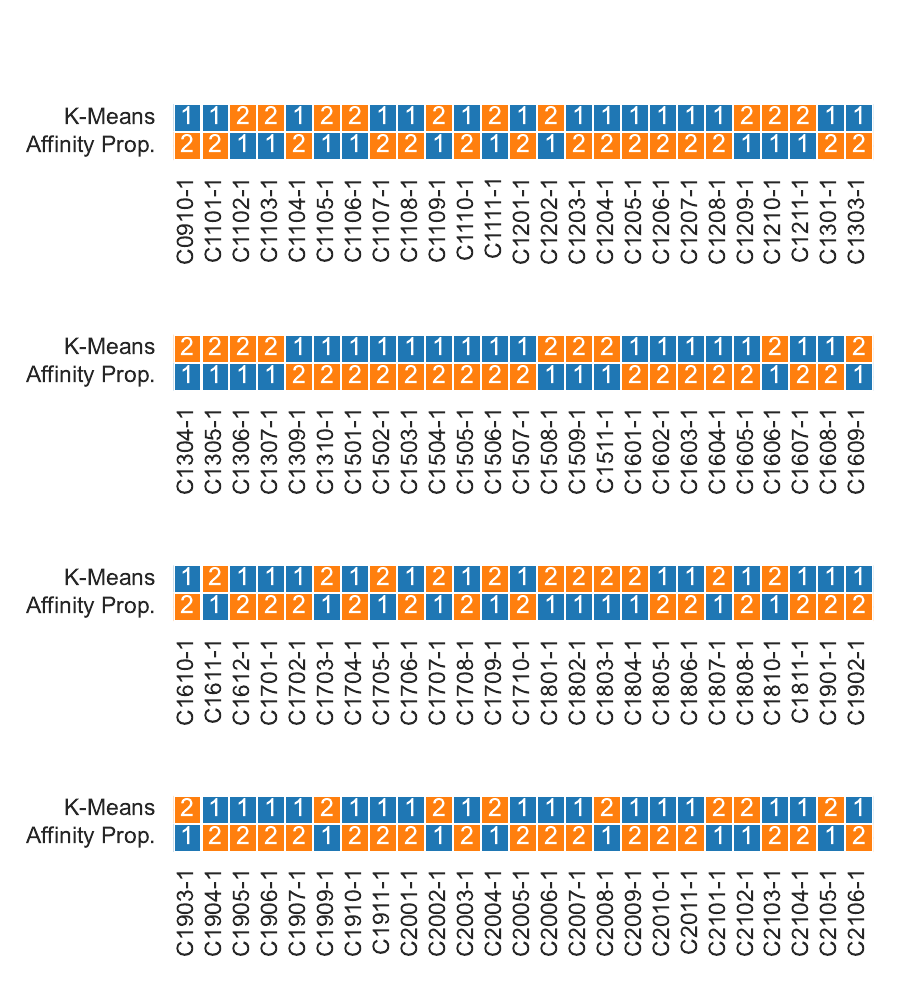}
        \caption{}\label{figure:cluster-partitioning-C5-b}
    \end{subfigure}
	\caption{(a) Two clusters of axial profile shapes identified by AP for the C5 assembly;  (b) $k$-means and AP methods assign different indices to these two clusters, but partitioning is consistent between them.
	}
	\label{figure:cluster-partitioning-C5}
\end{figure}

\begin{figure}[!htb]
	\centering
    \begin{subfigure}{0.575\textwidth}
        \centering
        \includegraphics[width=1\linewidth]{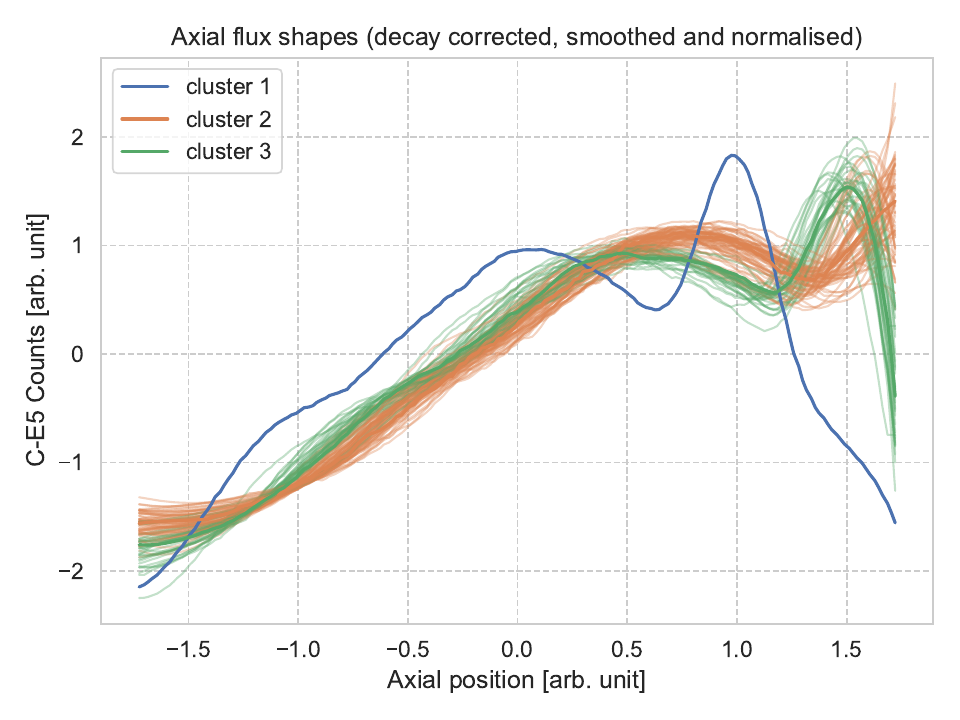}
        \caption{}\label{figure:cluster-partitioning-E5-a}
    \end{subfigure}
    \begin{subfigure}{0.41\textwidth}
        \centering
        \includegraphics[width=1\linewidth]{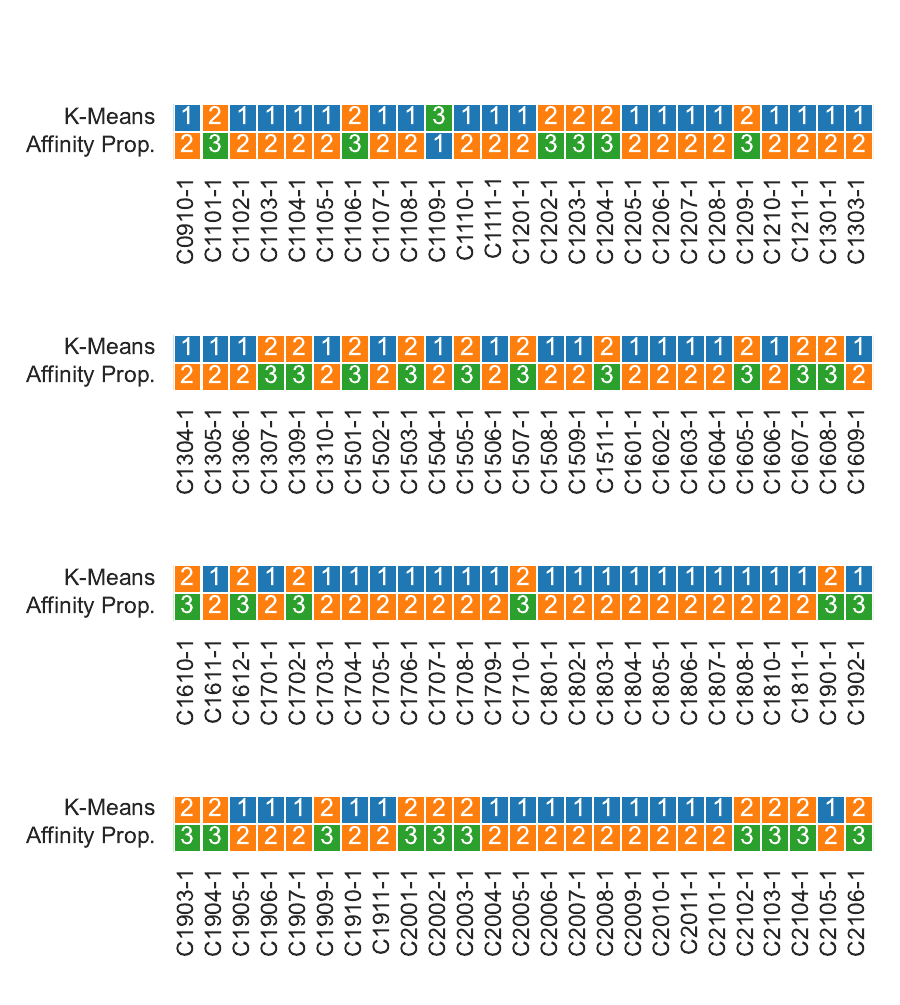}
        \caption{}\label{figure:cluster-partitioning-E5-b}
    \end{subfigure}
	\caption{(a) Three clusters of the axial profile shapes identified by AP for assembly E5; (b) Both clustering methods assign different indices to these clusters, but the partitioning is consistent between them, except for cycle C1902-1.
	}
	\label{figure:cluster-partitioning-E5}
\end{figure}

The clustering analysis was performed as described in Section~\ref{section:methodologies-clustering}. Values of hyperparameters that led to consistent cluster partitioning were $k=2  \text{ to } 3$ in $k$-means and $\text{``sample preference''} = -120$ (for C5, E7 and G5) and $-105$ (for C7, E5 and G7) in AP. As a result, two to three clusters were identified by both methods, as reported in Table~\ref{table:similarity-scores} and illustrated by Figures~\ref{figure:cluster-partitioning-C5} and \ref{figure:cluster-partitioning-E5} for core positions C5 and E5, respectively. In these figures, plots on the left-hand side contain $z$-score  normalized historical axial profiles, coloured according to the clusters they belong to, and thick lines indicating the centroids of corresponding clusters; pictures on the right-hand side report cluster indices attributed by both utilized clustering methods for the historical cycles (i.e. from cycle C0910-1 to cycle C2106-1).  A visual inspection of the axial profile shape plots for different core positions revealed that in the case of C7 and G7, one of three clusters corresponds to an evident outlier (as can be clearly seen in Figure~\ref{figure:cluster-partitioning-E5-a}), whereas a similar outlier, observed in the assembly G5 set, was not identified for the selected range of hyperparameters.

\begin{table}[!htb]
	\centering
	\caption{Number of clusters, cluster sizes and the similarity scores for the six CAs.}
	\label{table:similarity-scores}
        \footnotesize
        \begin{tabular}{lcccccc}
        \toprule 
        Parameter \textbackslash{} Core Position  & C5  & C7  & E5  & E7  & G5  & G7 \tabularnewline
        \midrule 
        \multicolumn{7}{l}{\textit{Number of clusters}} \tabularnewline
        \hspace{1em}$k$-means & 2  & 3  & 3  & 2  & 2  & 3 \tabularnewline
        \hspace{1em}Affinity Propagation & 2  & 3  & 3  & 2  & 2  & 3 
        \tabularnewline
        \multicolumn{7}{l}{\textit{Cluster sizes}} \tabularnewline
        \hspace{1em}$k$-means & 62+38  & 61+37+1  & 68+31+1  & 69+31  & 61+38$^*$  & 72+27+1 
        \tabularnewline
        \hspace{1em}Affinity Propagation & 62+38  & 61+37+1  & 67+32+1  & 69+31  & 61+38$^*$  & 72+27+1 \tabularnewline
        \multicolumn{7}{l}{\textit{Similarity Scores}}\tabularnewline
        \hspace{1em}Adjusted Rand Index  & 1.0  & 1.0  & 0.960  & 1.0  & 1.0  & 1.0 
        \tabularnewline
        \hspace{1em}Adjusted Mutual Information  & 1.0  & 1.0  & 0.926  & 1.0  & 1.0  & 1.0 \tabularnewline
        \hspace{1em}Normalized Mutual Information  & 1.0  & 1.0  & 0.928  & 1.0  & 1.0  & 1.0 \tabularnewline
        \bottomrule
        \multicolumn{7}{l}{$^*$Cycle C1501-1 does not contain data for G5}  \tabularnewline
        \end{tabular}
\end{table}

Once the number of clusters is identified, it is of interest to check if the cluster partitioning is done consistently by the two clustering analysis methods. An example of cluster partitioning is given in Figures~\ref{figure:cluster-partitioning-C5-b} and \ref{figure:cluster-partitioning-E5-b}. It can be seen in this example, that both methods partition the set of flux shapes from the historical cycles in the same way, although they assign different indices to the clusters. The cluster attribution consistency check can be done automatically by calculating indices ARI, AMI and NMI, which quantify the similarity between pairs of clusters and take different cluster indexing into account. Results of cluster similarity calculations are given in Table~\ref{table:similarity-scores} and indicate that the partitioning yielded by two clustering algorithms is exactly the same for all CAs except for E5. This minor mismatch (only one cycle, namely C1902-1, was attributed to different clusters) can be explained by a less prominent dissimilarity of measured flux shapes in E5 as compared to other positions.

\begin{figure}[htb]
	\centering
	\includegraphics[clip, trim=2.0cm 1.0cm 3.6cm 0.2cm, height=0.3\textwidth]{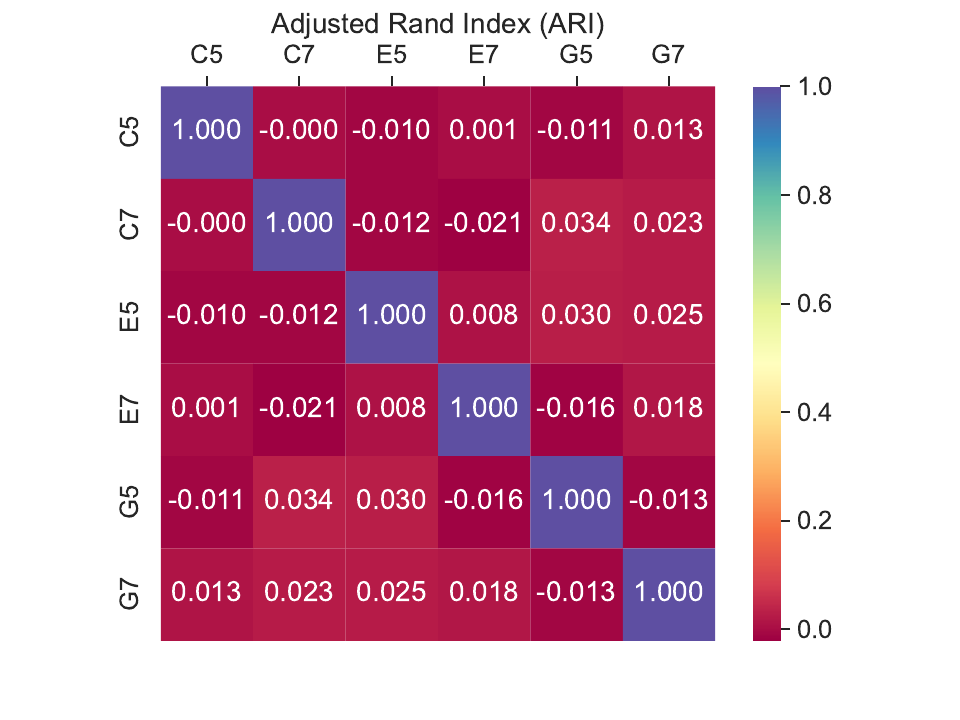}
	\includegraphics[clip, trim=2.7cm 1.0cm 3.6cm 0.2cm, height=0.3\textwidth]{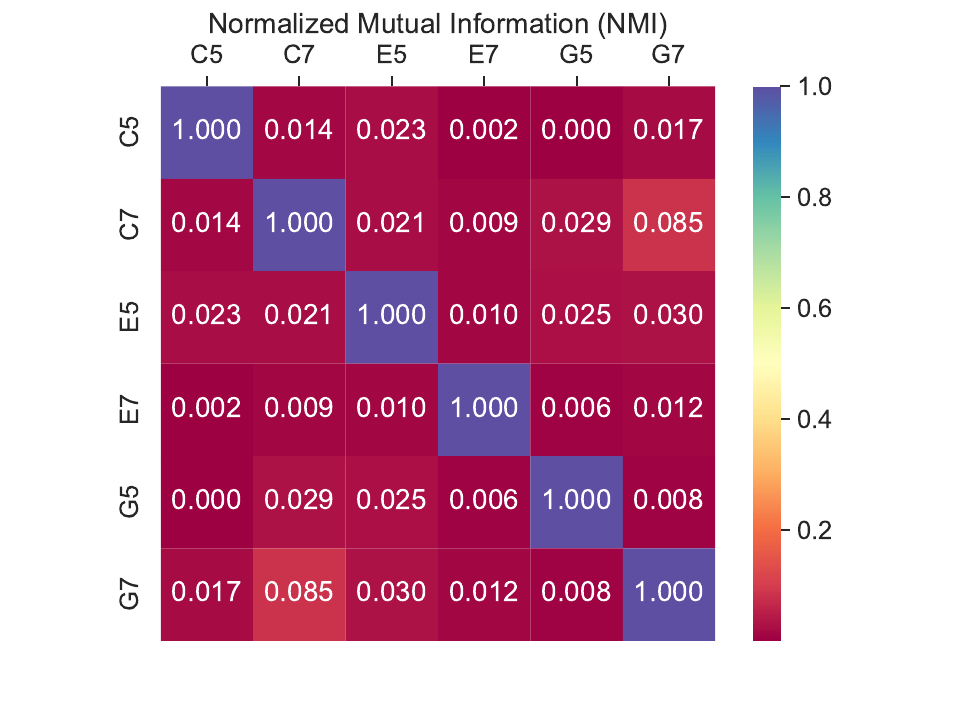}
	\includegraphics[clip, trim=2.7cm 1.0cm 2.0cm 0.2cm, height=0.3\textwidth]{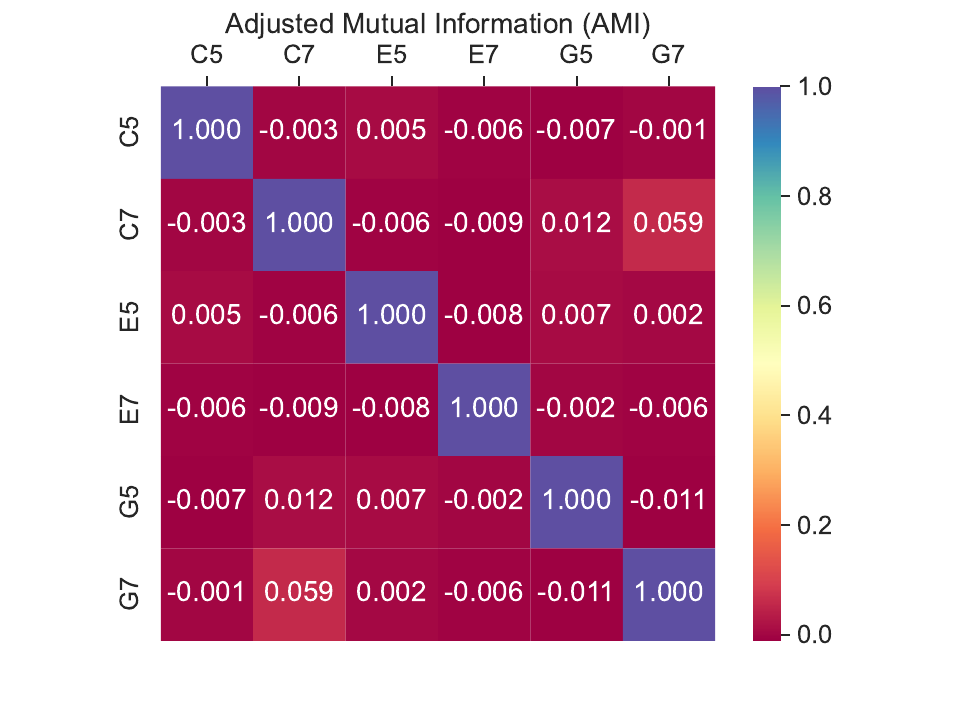}
	\caption{Similarity between clustering of CAs over historical cycles.}
	\label{figure:similarity-between-control-clustering}
\end{figure}

Indices ARI, AMI and NMI can also be applied to study the similarity between pairs of CAs in different core positions for historical cycles. A cluster correlation may (but does not have to) be an indication that clusters originate from some unknown external factor capable of affecting several CAs simultaneously. Results of this analysis are presented in Figure~\ref{figure:similarity-between-control-clustering} and imply that there is no or very little correlation between attributions of different assemblies to a cluster. One may therefore cautiously presume that the reason for the cluster appearance may be of a stochastic nature, such as an inadvertent misplacement of the blade containing a copper wire during the measurement. 

\begin{figure}[htb]
	\centering
	\includegraphics[width=0.995\textwidth]{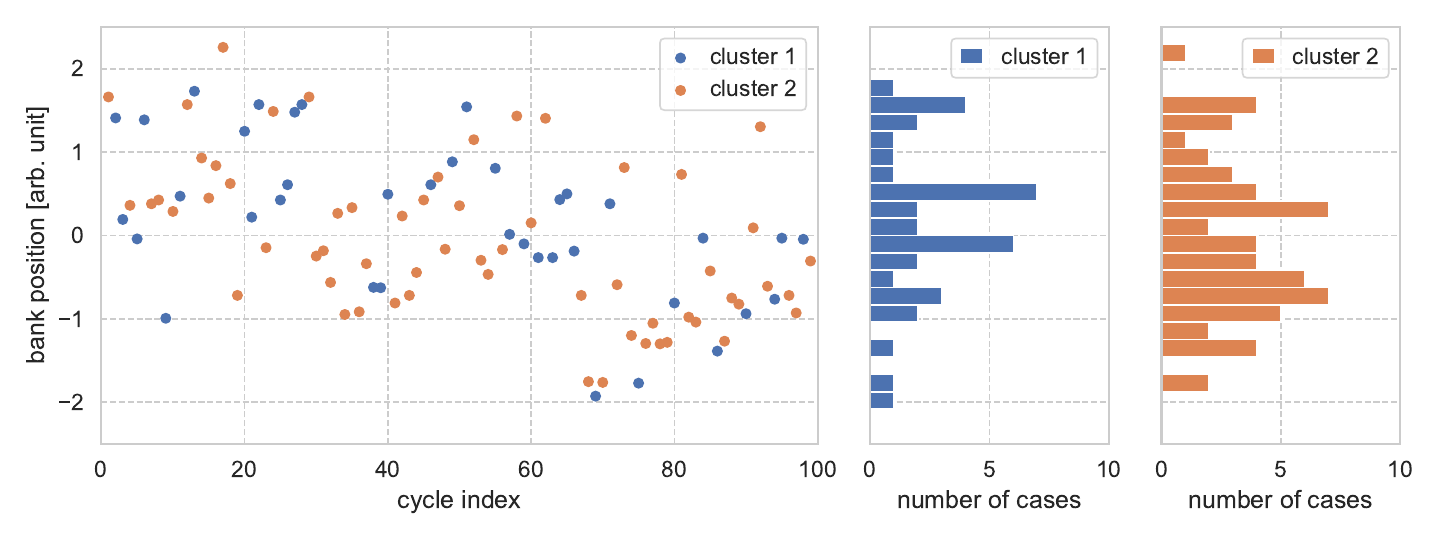}
    \caption{Distribution of the normalized control bank positions over cycles (in historical order) for clusters as identified by the $k$-means algorithm for core position C5.}
	\label{figure:control-bank-positions-for-cycles-C5}
\end{figure}

\begin{figure}[htb]
	\centering
    \includegraphics[width=0.995\textwidth]{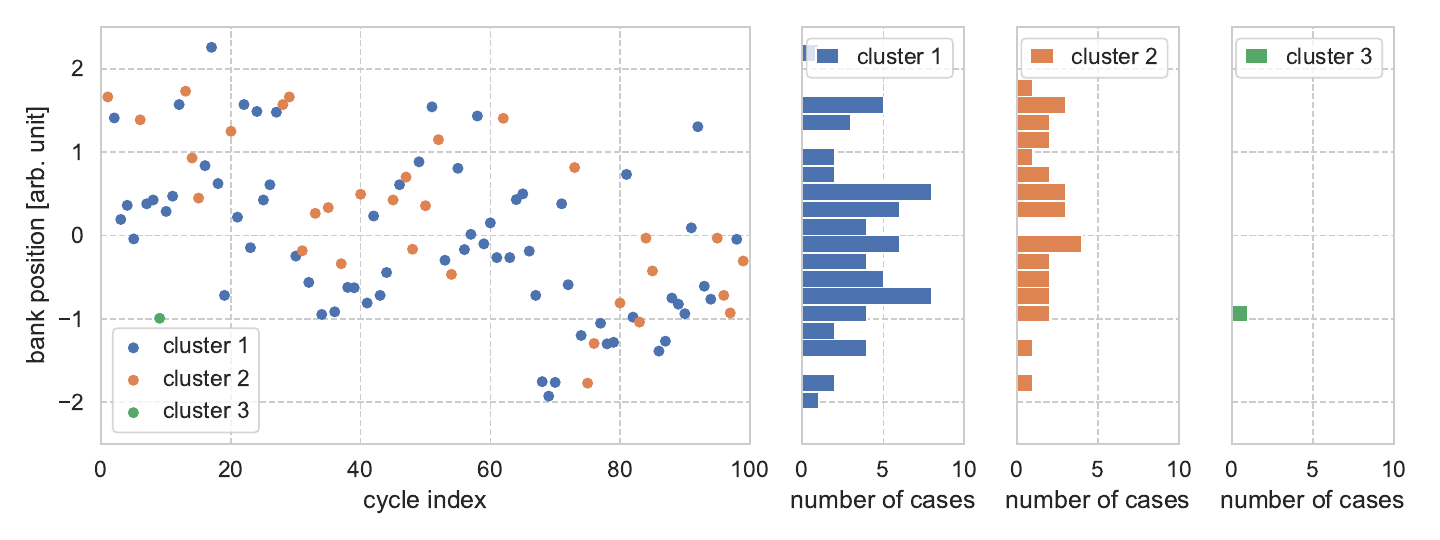}
    \caption{Distribution of the normalized control bank positions over cycles (in historical order) for clusters as identified by the $k$-means algorithm for core position E5.}
	\label{figure:control-bank-positions-for-cycles-E5}
\end{figure}

This presumption can further be supported by results of our analysis of the cluster distribution over historical cycles and control bank positions, which indicated that the clusters are distributed randomly over essentially the same range of these two parameters. This behaviour is illustrated by Figures~\ref{figure:control-bank-positions-for-cycles-C5} and \ref{figure:control-bank-positions-for-cycles-E5}, in which the distribution of the control bank positions over cycles for clusters as identified by the $k$-means algorithm for assemblies C5 and E5, respectively, is presented. In these figures, the cycles are arranged in historical order, i.e.~in the same order as given in Figures~\ref{figure:cluster-partitioning-C5-b} and \ref{figure:cluster-partitioning-E5-b}. In Figures~\ref{figure:control-bank-positions-for-cycles-C5} and \ref{figure:control-bank-positions-for-cycles-E5}, no pattern of cluster attribution can be observed in terms of both bank position and historical cycle: points corresponding to different clusters are scattered in a random manner and cover almost the entire ranges of both variables. Similar results were obtained for other CAs and for both clustering methods.

\subsection{Results for Regression Analysis and UQ}
\label{section:results-and-discussion-regression-and-UQ}

\begin{figure}[!htb]
	\centering
	\begin{subfigure}{0.495\linewidth}
		\centering
		\includegraphics[clip, trim=0.35cm 0.0cm 0.7cm 0.7cm, width=0.975\linewidth]{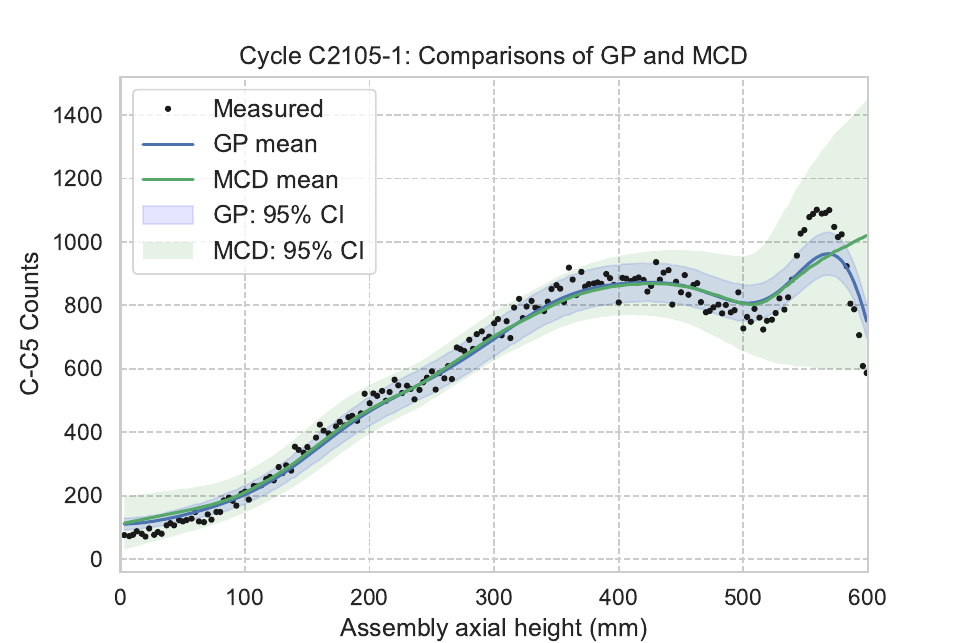}
		\caption{}
        \label{figure:flux-prediction-improvement-a}
	\end{subfigure}
	\begin{subfigure}{0.495\linewidth}
		\centering
		\includegraphics[clip, trim=0.35cm 0.0cm 0.7cm 0.7cm, width=0.975\linewidth]{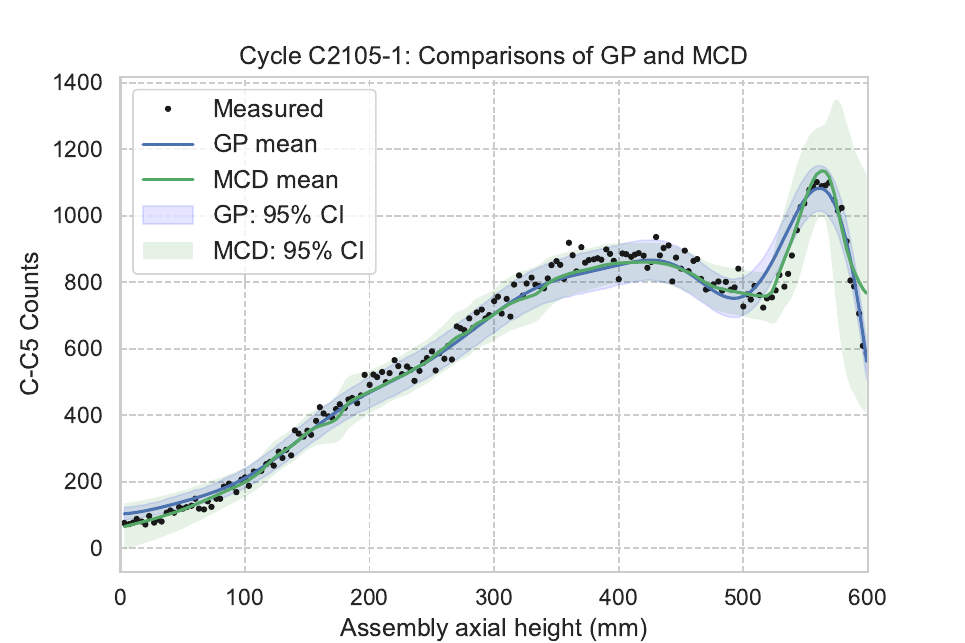}
		\caption{}\label{figure:flux-prediction-improvement-b}
	\end{subfigure}
	\begin{subfigure}{0.495\linewidth}
		\centering
		\includegraphics[clip, trim=0.35cm 0.0cm 0.7cm 0.7cm, width=0.975\linewidth]{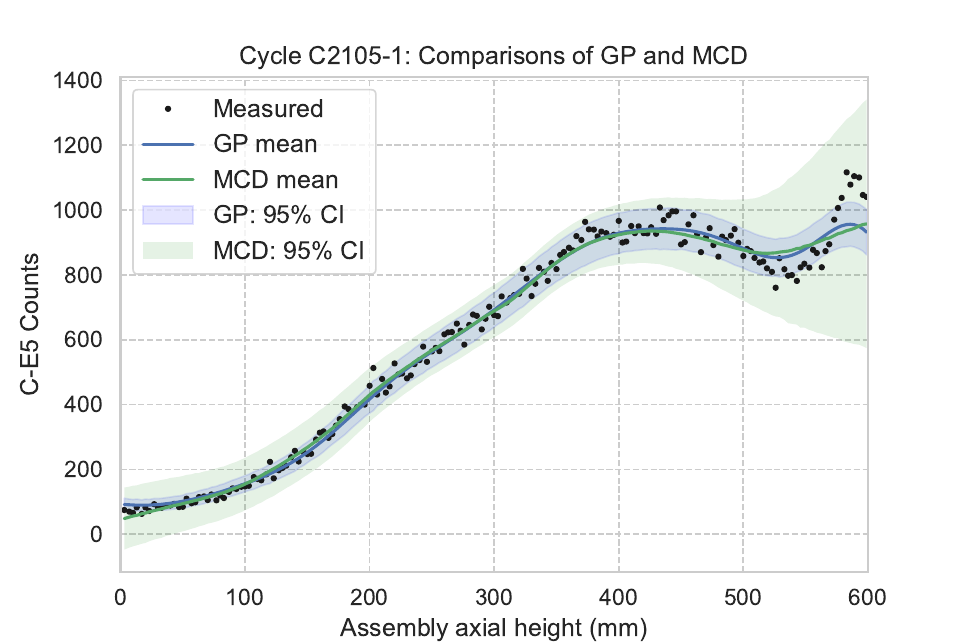}
		\caption{}\label{figure:flux-prediction-improvement-c}
	\end{subfigure}
	\begin{subfigure}{0.495\linewidth}
		\centering
		\includegraphics[clip, trim=0.35cm 0.0cm 0.7cm 0.7cm, width=0.975\linewidth]{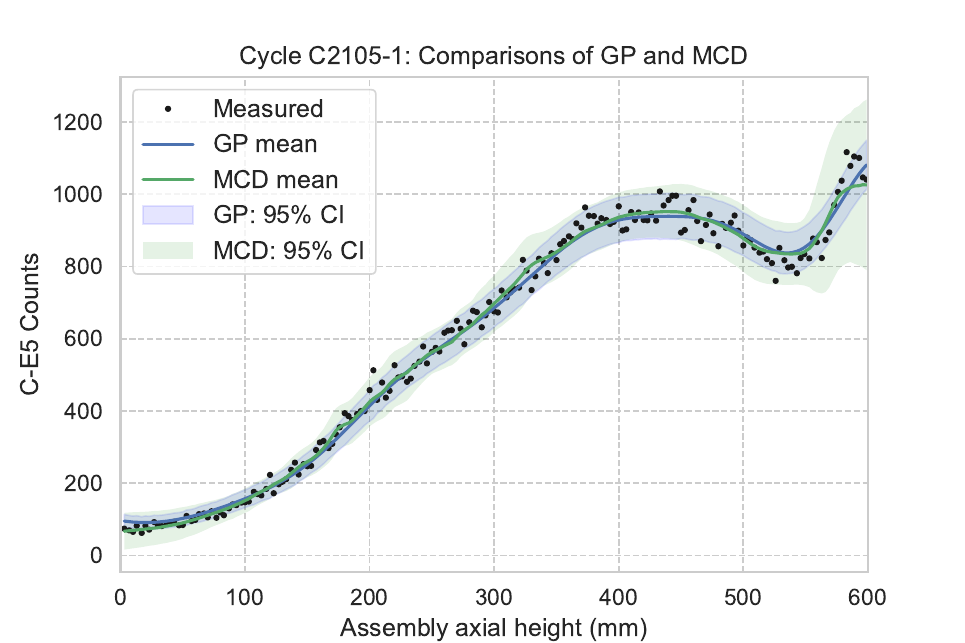}
		\caption{}\label{figure:flux-prediction-improvement-d}
	\end{subfigure}
	\caption{Predictions by GP and DNN without (left) and with (right) cluster partitioning.}
	\label{figure:flux-prediction-improvement}
\end{figure}

As the next step, separate GP and DNN models were trained  for each cluster of each CA. The predictions from these models are then compared with GP/DNN models trained using data without clustering, i.e.~with all the clusters together. Analysis of the obtained results revealed that, as a rule, after the separation to clusters, predicted flux profiles match the measurements better. This holds, in particular, for the peak in the vicinity of the coupling piece at the top of the CAs. Figure~\ref{figure:flux-prediction-improvement} illustrates the improvement of the axial flux profile prediction in cycle C2105-1 of the C5 and E5 assemblies, without (left) and with (right) the clustering procedure. In this figure, the uncertainties in predictions, shown by the shaded area, represent the \SI{95}{\percent} Confidence Interval (CI) around the mean indicated by lines. One may observe in Figures~\ref{figure:flux-prediction-improvement-b} and \ref{figure:flux-prediction-improvement-d} that the mean predictions by both GP and DNN are close and represent the measured values fairly well. This closeness of GP and DNN predictions is an important outcome of this study: given that the exact (reference) shapes are unknown, it increases the trustworthiness of the constructed models. 

Furthermore, after applying cluster separation, the uncertainty band envelops the experimental data more tightly. DNN uncertainty around the coupling piece reduces significantly but still remains bigger than GP uncertainty. Results from the other cycles and other CAs have demonstrated a similar behaviour. Reasons for the increased DNN uncertainty require additional studies. One possible reason is related to the sources of uncertainties in GP and DNN models, and how the output uncertainties are obtained: GP prediction uncertainty mainly reflect the noises in the data as well as data coverage, while DNN prediction uncertainty may also be notably influenced by model architecture (the model being too simple/complex) and randomness in training (initialization of parameters, gradient descent convergence, hyperparameter tuning, etc.).

\begin{figure}[!htb]
	\centering
    \begin{subfigure}{1\linewidth}
		\centering
		\includegraphics[clip, trim=0.0cm 0.0cm 0.0cm 0.3cm, width=0.99\textwidth]{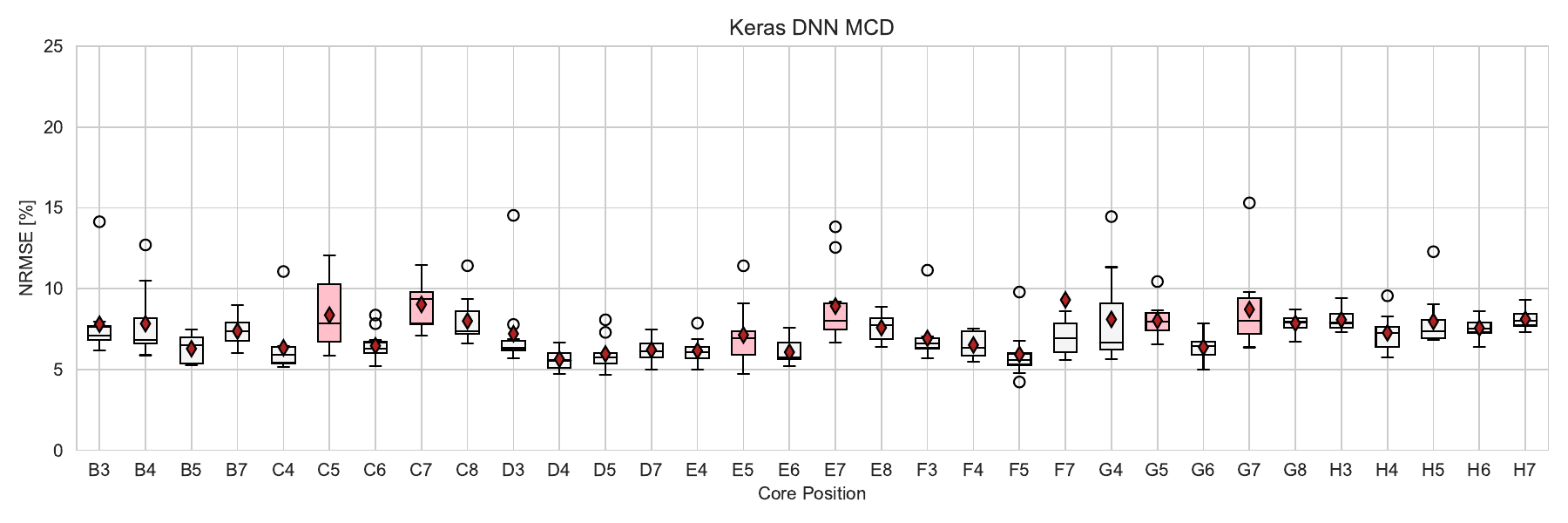}
		\caption{}\label{figure:box-plot-for-assemblies-gp-mcd-a}
	\end{subfigure}
 	\begin{subfigure}{1\linewidth}
		\centering
		\includegraphics[clip, trim=0.0cm 0.3cm 0.0cm 0.3cm, width=0.99\textwidth]{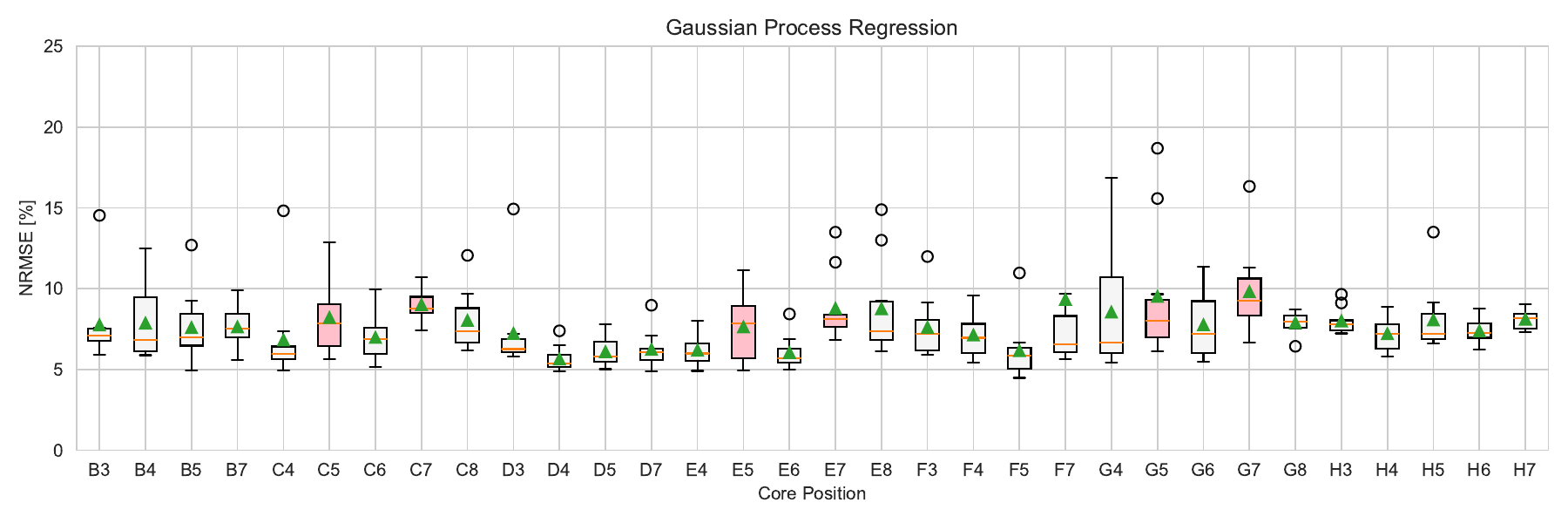}
		\caption{}\label{figure:box-plot-for-assemblies-gp-mcd-b}
	\end{subfigure}
 	\begin{subfigure}{1\linewidth}
		\centering
  \includegraphics[clip, trim=0.0cm 0.3cm 0.0cm 0.3cm, width=0.99\textwidth]{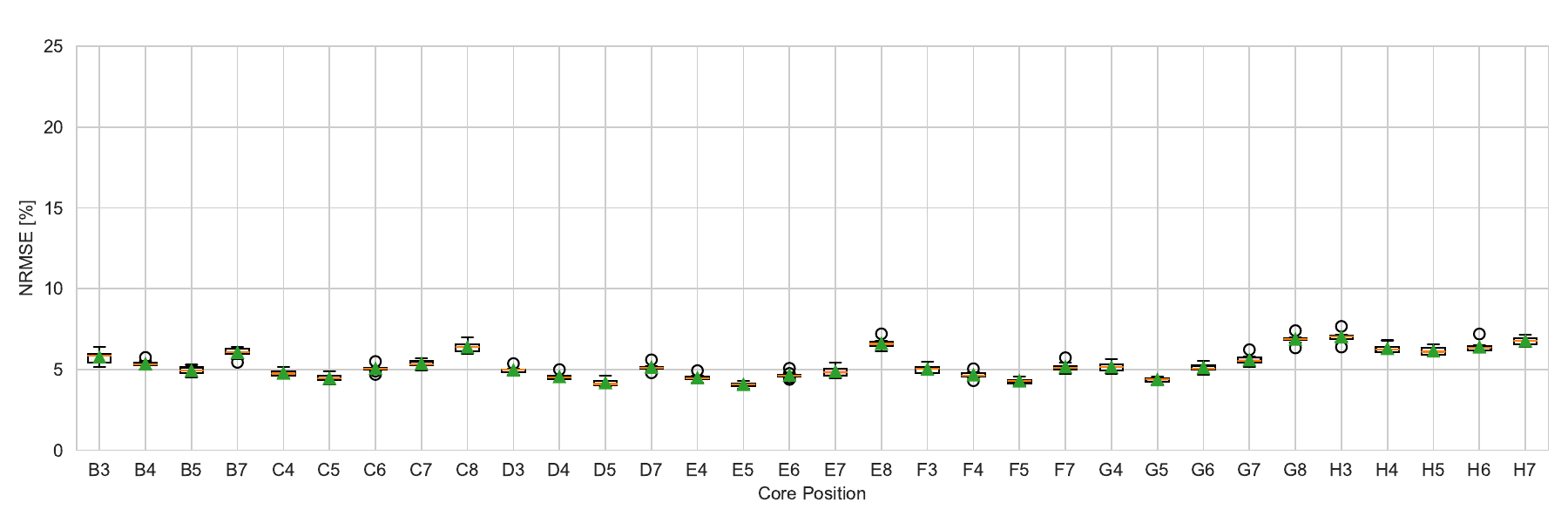}
		\caption{}\label{figure:box-plot-for-assemblies-gp-mcd-c}
	\end{subfigure}
	\caption{Percentage NRMSEs of thermal flux shape prediction for FA (white) and CA (pink) with cluster partitioning, using both DNN (a) and GP (b) ML models. For comparison, (c) provides an estimation of the error due to statistical noise, which is consistent with the approach used in (a) and (b).}
	\label{figure:box-plot-for-assemblies-gp-mcd}
\end{figure}

The overall performance of the enhanced ML models can be characterized by NRMSE box plots similar to those presented in Figure~\ref{figure:Percentage-Normalized-Root}.
%
%
The NRMSE distributions for DNN and GP over ten operational cycles, with clustering analysis considered, are represented by boxplots in Figure~\ref{figure:box-plot-for-assemblies-gp-mcd}, for all 32 assemblies. Upon comparing these results to the non-cluster case in Figure~\ref{figure:Percentage-Normalized-Root}, a significant improvement in all the follower NRMSEs can be clearly noticed.
Furthermore, similar trends in error distribution, with small cycle-to-cycle variations are displayed by both regression methods.
As may be observed in Figure~\ref{figure:Percentage-Normalized-Root}, the FAs' error lies within a range of \SI{5}{\percent} to \SI{10}{\percent} while the CAs showed higher errors within a wider range of \SI{6}{\percent} to \SI{23}{\percent}. Applying the clustering analysis for the CAs resulted in comparatively shorter and lower-placed box plots, indicating a better prediction accuracy for both DNN and GP. With the exception of a few outliers, the error has been reduced to less than \SI{11}{\percent} for both DNN and GP. The prediction accuracy of both DNN and GP seems to be somewhat similar across all the follower assemblies.

It would be valuable to qualify the overall performance of the final set of the constructed fuel- and control assembly models. This may be achieved, for example, by comparing the prediction accuracy to the statistical errors for the same set of test cycles. To this end, the NRMSE due to the noise was calculated in a consistent manner 
and results are presented in Figure~\ref{figure:box-plot-for-assemblies-gp-mcd-c}. As one may observe, the statistical error varies between \SI{4}{\percent} and \SI{7}{\percent} depending on the core position with little variation between cycles, which indicates 
consistent measurement conditions. Furthermore, analysis of data presented in Figure~\ref{figure:box-plot-for-assemblies-gp-mcd} allows assessing DNN and GP approximation errors (in terms of NRMSEs). The analysis yields the following very rough estimation of the approximation error: \SI{3}{\percent} to \SI{5}{\percent} for FAs and \SI{6}{\percent} to \SI{7}{\percent} for CAs, depending on the core position.   

As a final remark, we would like to mention that splitting the training dataset to clusters resulted in, besides an improved prediction accuracy, a significant acceleration in the ML model training time, especially in the case of GP, for which the training time for the ensemble of CA is reduced by a factor of three.

\section{Conclusions}
\label{section:conclusions}

In this work, we developed a procedure that combines unsupervised and supervised ML to predict the assembly axial neutron flux profiles in the SAFARI-1 research reactor, trained by measurement data from multiple historical cycles. The motivation was to improve the accuracy of ML model predictions for the SAFARI-1 control follower assemblies when compared with the measurement data, which was found to be noticeably lower than the accuracy for fuel assemblies in previous works. With unsupervised ML, clustering analysis using $k$-means and affinity propagation algorithms was implemented to identify the clusters in the measured axial neutron flux profiles. Then, regression-based supervised ML models using DNN and GP were trained for different clusters, with their approximation/prediction uncertainties quantified. 

It was demonstrated that both DNN and GP models had much better performance after clustering the dataset for the control follower assemblies in terms of flux shape and uncertainty prediction. An additional benefit of using clustered data is an improved computational performance of GP training and reconstruction. The quantified uncertainties in GP and DNN predictions exhibited, however, large differences in certain assembly axial locations. Future work will be devoted to explaining the smaller uncertainties observed in GP-based ML models.


%
%
%
%
%
%
\section*{CRediT authorship contribution statement}

\textbf{L.~E.~Moloko}: Methodology, Software, Formal analysis, Investigation, Writing -- original draft,
Writing -- review \& editing, Visualization.
\textbf{P.~M.~Bokov}: Conceptualization, Methodology, Software, Validation, Formal analysis, Investigation, Writing - original draft, Writing -- review \& editing, Visualization.
\textbf{X.~Wu}: Methodology, Writing -- original draft, Writing -- review \& editing.
\textbf{K. N. Ivanov}: Conceptualization, Writing -- review.
\section*{Declaration of Competing Interest}

The authors declare that they have no known competing financial
interests or personal relationships that could have appeared
to influence the work reported in this paper.
\section*{Acknowledgement}
The authors are grateful to Ms. Hantie Labuschagne for proofreading the manuscript.
%
\bibliography{./bibliography.bib}

\appendix
\section{Definitions of Selected Cluster Similarity Measures}
\label{sec:Appendix-A}
 
%
We begin by introducing notations, used in this section, and providing necessary definitions. Let $\mathcal{S}$ be a set of $N$ objects, then a (partitional) clustering $\mathcal{C}$ is a set $\{C_1,\ldots, C_r\}$ of non-empty disjoint subsets of $\mathcal{S}$ such that their union equals $\mathcal{S}$. The set of all clusterings of $\mathcal{S}$ is denoted by $\mathcal{P}(\mathcal{S})$. Let $C' = \{C'_1,\ldots,C'_s\} \in \mathcal{P}(\mathcal{S})$ denote another clustering of $\mathcal{S}$. Furthermore, $|\cdot|$ denotes the cardinality (size) of a set, hence $|\mathcal{S}|=N$, $|\mathcal{C}|=r$ and $|\mathcal{C}'|=s$. 

Numerous measures for comparing clusterings have been proposed.  According to a taxonomy introduced in \cite{Vinh210Information}, the cluster similarity can be grouped into the class of the set-matching-based measures, pair-counting-based measures, and information-theoretic measures. Here we will focus on the last two, since they are utilized in our study. 

%
The pair-counting-based measures are constructed by counting pairs of objects on which two clusterings $\mathcal{C}$ and $\mathcal{C}'$ agree or disagree. Let $n_{11}$ be the number of pairs that are in the same cluster in both $\mathcal{C}$ and $\mathcal{C}'$, $n_{00}$ is the number of pairs that are in different clusters in both $\mathcal{C}$ and $\mathcal{C}'$, $n_{01}$ is the number of pairs that are in the same cluster in $\mathcal{C}$ but in different clusters in $\mathcal{C}'$ and $n_{10}$ is the number of pairs that are in different clusters in $\mathcal{C}$ but in the same cluster in $\mathcal{C'}$. 
For two clusterings $\mathcal{C}$ and $\mathcal{C}'$, the Rand Index (RI) or Rand measure (named after William M. Rand) is defined as \cite{Rand1971Objective}
\begin{equation}
\mathrm{RI}(\mathcal{C},\mathcal{C}') = 
\dfrac{n_{11}+n_{00}}{n_{00} + n_{01} + n_{10} + n_{11}} = \dfrac{2(n_{11}+n_{00})}{N(N-1)}.
\label{eq:Rand-Index}
\end{equation}
Nominally, the RI ranges from 0 (in a case where no pairs are classified in the same way under both clusterings) to 1 (for identical clusterings). The RI has been mostly used in it adjusted form, known as the Adjusted Rand Index (ARI), defined as \cite{Hubert1985Comparing}:
\begin{equation}
\mathrm{ARI}(\mathcal{C},\mathcal{C'}) = 
\dfrac{2(n_{00}n_{11}-n_{01}n_{10})}{(n_{00}+n_{01})(n_{01}+n_{11})+(n_{00}+n_{10})(n_{10}+n_{11})}.
\label{eq:Adjusted-Rand-Index}
\end{equation}
ARI has an expected value zero for independent clusterings (under the generalized hypergeometric distribution assumption for randomness) and is bounded by one (for identical clusterings). \cite{Vinh210Information}. Note, that some pairs of clusterings may result in negative index values \cite[and references therein]{Wagner2007Comparing}.

%
The information-theory-based measures are based on concepts from the information theory: entropy, joint entropy, conditional entropy and mutual information \cite{Vinh210Information}. In the context of clustering, the entropy and the mutual information are defined in the following way. 
Let us assume that all elements of $\mathcal{S}$ have the same probability of being picked and choosing an element of $\mathcal{S}$ at random, the probability that this element is in cluster $C_i \in \mathcal{C}$ is $p_i = |C_i|/N$ \cite{Wagner2007Comparing}. Then the entropy associated with clustering $\mathcal{C}$ is defined as 
\begin{equation}
    H(\mathcal{C}) = - \sum_{i=1}^{r}p_i\log_2 p_i,
\end{equation}
The mutual information between two clusterings $\mathcal{C}$ and $\mathcal{C}'$ is defined as
\begin{equation}
    I(\mathcal{C}, \mathcal{C}') = \sum_{i=1}^{r} \sum_{j=1}^{s} p_{ij} \log_2 \dfrac{p_{ij}}{p_i p_j},
\end{equation}
where $p_{ij}$ denotes the probability that an element belongs to cluster $C_i \in \mathcal{C}$ and to cluster $C'_j \in \mathcal{C}'$ given by $p_{ij} = |C_i \cap C'_j |/N$. The NMI and AMI are defined as  
\begin{equation}
    \mathrm{NMI}(\mathcal{C}, \mathcal{C}') = 
    \dfrac{I(\mathcal{C}, \mathcal{C}')}{\mathrm{mean}[H(\mathcal{C}), H(\mathcal{C}')]}
    \label{eq:Normalized-Mutual-Information}
\end{equation}
and
\begin{equation}
    \mathrm{AMI}(\mathcal{C}, \mathcal{C}') = 
    \dfrac{I(\mathcal{C}, \mathcal{C}') - \mathbb{E}[I(\mathcal{C}, \mathcal{C}')]}{\mathrm{mean}[H(\mathcal{C}), H(\mathcal{C}')] - \mathbb{E}[I(\mathcal{C}, \mathcal{C}')]},
    \label{eq:Adjusted-Mutual-Information}
\end{equation}
respectively, where ``$\mathrm{mean}[H(\mathcal{C}), H(\mathcal{C}')]$'' denotes some generalized mean of the entropies of each clustering \cite{JMLR:v12:pedregosa11a}. Various definitions of generalized means were proposed in the literature, for instance the arithmetic one: $\left[H(\mathcal{C}) + H(\mathcal{C}')\right]/2$, the geometric one: $\sqrt{H(\mathcal{C}) H(\mathcal{C}')}$ and others. More details on information-theoretic measures, as well on their different adjusted-for-chance and normalized forms, may be found in \cite{Vinh210Information}.

\end{document}